 \documentclass[pmlr,twocolumn,10pt]{jmlr} 

\usepackage{booktabs}
\usepackage{multirow}
\usepackage{makecell}
\usepackage{pifont}
\usepackage[capitalize]{cleveref}
\usepackage[per-mode=symbol]{siunitx}
\usepackage{csvsimple}
\newcommand{\cmark}{\ding{51}}
\newcommand{\xmark}{\ding{55}}
\DeclareMathOperator{\cov}{cov}

\DeclareMathOperator {\diag}{diag}

\usepackage{siunitx}

\usepackage{xspace}
\newcommand*{\modelPresp}{\texttt{P-Resp}\xspace}
\newcommand*{\modelPidr}{\texttt{P-IDR}\xspace}
\newcommand*{\modelGPresp}{\texttt{GP-Resp}\xspace}
\newcommand*{\modelGPlfm}{\texttt{GP-LFM}\xspace}
\newcommand*{\modelGPconv}{\texttt{GP-Conv}\xspace}

\newcommand*{\abstime}{\ensuremath{\tau}}
\newcommand*{\vecabstime}{\ensuremath{\boldsymbol{\tau}}}
\newcommand*{\reltime}{\ensuremath{\Delta\tau}}
\newcommand*{\vecreltime}{\ensuremath{\Delta\boldsymbol{\tau}}}

\theorembodyfont{\upshape}
\theoremheaderfont{\scshape}
\theorempostheader{:}
\theoremsep{\newline}

\jmlrvolume{225}
\jmlryear{2023}
\jmlrsubmitted{LEAVE UNSET}
\jmlrpublished{LEAVE UNSET}
\jmlrworkshop{Machine Learning for Health (ML4H) 2023} 

 \title[Nonparametric modeling of blood glucose dynamics]{Nonparametric modeling of the composite effect of multiple nutrients on blood glucose dynamics}

\author{%
\Name{Arina Odnoblyudova}\textsuperscript{1}\Email{arina.odnoblyudova@aalto.fi}\\
\Name{{\c{C}}a{\u{g}}lar Hizli}\textsuperscript{1} \Email{caglar.hizli@aalto.fi}\\
\Name{ST John}\textsuperscript{1} \Email{ti.john@aalto.fi}\\
\Name{Andrea Cognolato}\textsuperscript{1} \Email{andrecogno@hotmail.it}\\
\Name{Anne Juuti}\textsuperscript{2} \Email{anne.juuti@hus.fi}\\
\Name{Simo Särkkä}\textsuperscript{1} \Email{simo.sarkka@aalto.fi}\\
\Name{Kirsi Pietil{\"a}inen}\textsuperscript{2} \Email{kirsi.pietilainen@helsinki.fi}\\
\Name{Pekka Marttinen}\textsuperscript{1} \Email{pekka.marttinen@aalto.fi}\\
\addr \textsuperscript{1} Aalto University, Finland \\
\addr \textsuperscript{2} University of Helsinki, Finland
}

\begin{document}

\maketitle

\begin{abstract}
In biomedical applications it is often necessary to estimate a physiological response to a treatment consisting of multiple components, and learn the separate effects of the components in addition to the joint effect. Here, we extend existing probabilistic nonparametric approaches to explicitly address this problem. We also develop a new convolution-based model for composite treatment--response curves that is more biologically interpretable. We validate our models by estimating the impact of carbohydrate and fat in meals on blood glucose. By differentiating treatment components, incorporating their dosages, and sharing statistical information across patients via a hierarchical multi-output Gaussian process, our method improves prediction accuracy over existing approaches, and allows us to interpret the different effects of carbohydrates and fat on the overall glucose response.

\end{abstract}

\section{Introduction}
\label{sec:intro}
External factors have a profound impact on biological systems. One notable example is the effect of nutrients obtained through food, which can alter blood glucose levels in conjunction with various other factors \citep{albers2017personalized, balakrishnan2014personalized}. Comprehending how these factors shape a patient's physiology is clinically important and crucial for personalized therapy, for example to maintain the blood glucose level in a desired healthy range \citep{lunceford2002estimation, murphy2007customizing, murphy2007developing,silva2016observational}. A high-impact application of glucose modeling is diabetes management, where accurate modeling aids in crafting personalized treatment plans. This requires comprehending the personalized impact of different components of diet and other factors, in order to regulate them appropriately.

Treatment--response curves (TRCs) capture the dynamics of temporal physiological signals when subjected to various treatments \citep{robins1987graphical,robins2000marginal, gill2001causal,bang2005doubly}. TRCs are commonly represented by a continuous-time function $y$ modeled as the sum of the baseline trend $f_b$ and the responses $f_r$ to treatments with some dosages at given times \citep{trm_counterfactual_reasoning,eiv,pmlr-v56-Xu16,hizli2023causal}. In many real-world cases, treatments are composed of multiple components such that each component affects the response differently. Thus, we assume that the overall response $f_r$ depends on the addition or more complex combination of multiple functions $\{f_{rq}\}_{q=1}^Q$, corresponding to different treatment components indexed by $q$. This is illustrated in \figureref{fig:trcintro}, where we see how the presence and amount of a second treatment component (fat) changes the response caused by the first component (sugar). Correspondingly, the two goals in this setup are: (i) modeling of the overall treatment--response function $f_{r}$ to make personalized predictions for the physiological quantity $y$ under varying treatments, and (ii) evaluating the component functions $f_{rq}$ and their contributions to $f_{r}$.
\looseness-1

\begin{figure}[htbp]
\floatconts
  {fig:trcintro}
  {\caption{Illustrated glucose response under various treatment setups. \textbf{Left:} Baseline without any treatment. \textbf{Middle:} Response to one treatment component (10g sugar). \textbf{Right:} Response to two treatment components (10g sugar and 3g fat). Addition of fat increases the peak, but also delays it in time.\looseness-1}}
  {\includegraphics[width=1.0\linewidth, trim={0 0.15cm 0 0}, clip]{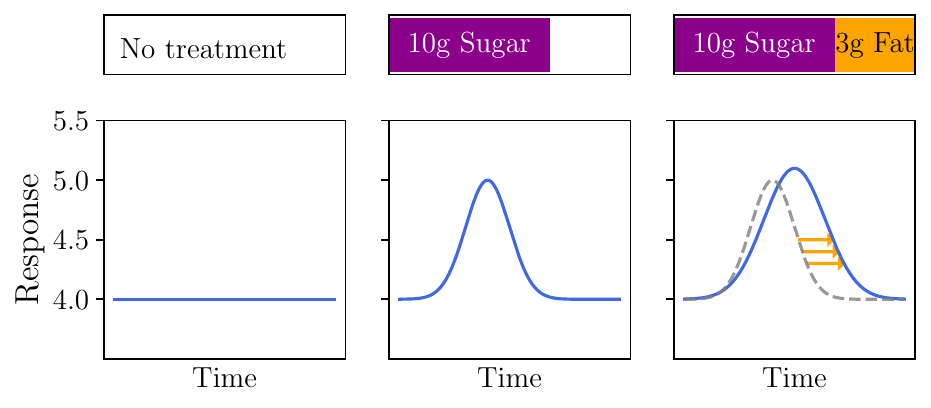}}
\end{figure}

Estimating response curves of composite treatments is challenging due to several characteristics of complex longitudinal healthcare data: measurements are often sparse, noisy, and irregularly sampled, with a varying number of time points across patients \citep{gustafson2003measurement, hall2008measurement}. Moreover, it is not clear how to model the effect of treatment components on the overall outcome in the case of composite treatments. In particular, whether these effects should be estimated separately or jointly with some kind of interaction between them. Existing works, like \citet{eiv,hizli2023causal}, only model a single overall response curve, which cannot address interactions between multiple treatment components. Yet another question is how to include treatment dosage information in the model. Some studies, e.g. \citet{pmlr-v56-Xu16, psem_gp_lfm_cheng}, do not include it at all. Modern modeling approaches need to address all these issues by being flexible and incorporating uncertainty  \citep{pml_healthcare_book, spiegelhalter1999introduction} to accurately characterize the biomedical observations.

To address these challenges, we investigate Bayesian parametric and nonparametric models for treatment--response curves. Usually, the nonparametric methods are beneficial due to their greater flexibility and ability to capture complex dynamics in longitudinal data  \citep{ferguson1973bayesian, muller2013bayesian, rodriguez2013nonparametric}. However, if domain experts have an accurate understanding of the desired TRC form, tailored parametric models may be efficient.

Our contributions can be summarized as follows:
\begin{enumerate}
        \item{We extend existing nonparametric models for TRCs to composite treatments, such that each treatment component's influence on the response curve becomes separately characterized. The models are further improved by incorporating treatment dosages using Multi-output GPs (MOGPs). This approach also facilitates a joint analysis across multiple patients.}
        \item{We develop a novel nonparametric TRC model based on a convolution of a Gaussian process (GP) that can capture the interaction between treatment components in a way that reflects the biological intuition, making it more interpretable.} 
        \item{We thoroughly compare the developed models with the existing work to verify their suitability to accurately predict the response of blood glucose to carbohydrate (carbs) and fat intake from meals. From the biological perspective, convolution-based model estimates the driving effect of carbohydrates and confirms the delayed effect of fat.}
\end{enumerate}
The experiments are done on a real-world blood-glucose dataset from the Helsinki University Hospital, which includes continuous blood glucose measurements and the treatment regimens of the patients, represented by meals eaten throughout a period of three days. The treatments are characterized by two main components: the amounts of carbohydrates and fat.

\begin{table*}[htbp]
\floatconts
  {tab:modelscomparison}%
  {\caption{Overview of compared methods. Our parametric models \modelPresp and \modelPidr are described in \appendixref{apd:second}. Our nonparametric models  \modelGPresp, \modelGPlfm and \modelGPconv are described in \sectionref{sec:npseptt,sec:nplfm,sec:npconv}, respectively.}}%
  {%
  \small
  \colorlet{good}{green!70!black}
  \colorlet{bad}{red!70!black}
  \newcommand*{\yes}{{\color{good}\cmark}}
  \newcommand*{\no}{{\color{bad}\xmark}}
  \newcommand*{\NA}{\textcolor{bad}{Not available}}
\begin{tabular}{lcccccc}\toprule
& Model & \makecell{Response \\ modeling} & \makecell{Dosage \\ dependence} & \makecell{Multiple \\ individuals} &  \makecell{Different \\ treatment components}\\
\midrule
    \parbox[t]{3mm}{\multirow{4}{*}{\rotatebox[origin=c]{90}{Parametric\vphantom{p}}}}
    & \citet{trm_counterfactual_reasoning} & Linear time-invariant & \yes & Hierarchical model  & \NA \\ 
    & \citet{eiv}                          & Bell curve     & \yes & Hierarchical model  & Shared response \\
    & \modelPresp  & Bell curve (independent)  & \yes & Hierarchical model & \color{good} Additive  \\
    & \modelPidr   & Bell curve (dependent)  & \yes & Hierarchical model & \color{good} Additive  \\
    \midrule
    \parbox[t]{3mm}{\multirow{6}{*}{\rotatebox[origin=c]{90}{Nonparametric}}}
    & \citet{pmlr-v56-Xu16}             & Dirichlet Process     & \no  & Hierarchical model  & \NA \\
    & \citet{hizli2023causal}           & Gaussian Process      & \yes & Multi-output GP    & \NA \\
    & \citet{psem_gp_lfm_cheng}         & Latent force model    & \no  & Hierarchical model  & \NA \\
    & \modelGPresp & Gaussian Process   & \yes & Multi-output GP & \color{good} Additive   \\
    & \modelGPlfm  & Latent force model & \yes & Multi-output GP & \color{good} Additive \\
    & \modelGPconv & Convolution        & \yes & Multi-output GP & \color{good}\bfseries Jointly   \\
\bottomrule
\end{tabular}
}
\end{table*}

\section{Related Work}
\label{sec:relatedwork}
We compare existing methods from the literature and the models we developed in \tableref{tab:modelscomparison}, based on the following dimensions: parametric vs.~nonparametric, type of the response function, the incorporation of treatment dosage in the model, the way of modeling data across all patients, and the possibility to distinguish the effects of different treatment components. The last dimension is the most relevant for the present study. When treatments are modeled jointly, it implies that their responses are coupled in some way, whereas additive treatments are modeled separately and simply summed up for the overall response curve. Descriptions of the developed parametric models are shown in \appendixref{apd:second}, and they serve as baselines for the more advanced methods.

\paragraph{Parametric modeling.} \citet{trm_counterfactual_reasoning} represent the TRCs as a linear time-invariant second-order dynamical system. The response function's structure is shared across multiple signals.
\citet{eiv} introduce a parametric method in which individualized treatment--response curves are represented as bell-shaped response functions, hierarchically sharing patients' information. The parameter specifying the magnitude of the bell-shaped curve is a linear combination of the dosages of various treatment components. The model accounts for errors in treatments' values and times.

\paragraph{Nonparametric modeling.} In \citet{pmlr-v56-Xu16}, the response consists of two main components: the baseline progression and additive effects of treatment--response functions, where individual treatment curves are parameterized as ``U''-shaped functions. Dirichlet Process Mixture prior is used to cluster the baseline progression and the treatment response
parameters keeping individual-specific variability. \citet{hizli2023causal} focus on the usage of causality in learning TRCs. The authors model all patients jointly with the help of multi-output GPs but do not differentiate between various treatment components. \cite{psem_gp_lfm_cheng} model TRCs using a GP-based Latent Force Model (LFM). This adds a mechanistic perspective to purely data-driven GPs. Moreover, they force effects to only act forward in time using causal time-marked kernels. The LFM approach is presented in \citet{gps_transcriptionalreg} and is illustrated with three case studies from computational biology, motion capture, and geostatistics. In the work by \citet{lfms_alvarez09a}, the latent modeling approach has been tested to discover the dynamics of transcriptional processes in the cell. 

\paragraph{Convolution modeling.}
Convolution is a mathematical operation that can be applied to a waveform to filter it in some specific way. In blood-glucose modeling, it can be used to adjust the form of the response function, however, the following approach has not been used in TRCs modeling problem yet. It has been applied in the context of GPs construction for temporal data by convolving some continuous white noise process, for example, in \citet{higdon2002space}. In \citet{boyle2004dependent} Gaussian processes are treated as white noise sources convolved with smoothing kernels which is equivalent to stimulating linear filters with Gaussian noise. Using this, authors extend GPs to handle multiple, coupled outputs. \citet{alvarez2010efficient} construct multioutput Mercer kernel with convolution processes. It allows the integration of prior information from physical models, such as ordinary differential equations, into the covariance function. However, the main focus of the paper is on efficient inference with convolved GPs.

\section{Methods}
\label{sec:methods}

The following section provides a comprehensive description of the devised models. We first formulate the high-level TRC model. From this general form we develop its parametric and nonparametric extensions and present how they differ from each other. We primarily focus on the nonparametric response models, which include one conventional GP model (\modelGPresp), a Latent Force Model (\modelGPlfm), and a newly developed Convolution Model (\modelGPconv).

\paragraph{Model formulation.} Here we present the general,  high-level model overview. Let the model of the physiological outcome for a single patient be described as:
\begin{equation}\label{eq:model}
    y(\vecabstime)=f^{b}(\vecabstime)+\sum_{j=1}^{M}f^{r}(\vecabstime,t_{j},\mathbf{m}_{j})+ \boldsymbol{\epsilon},
\end{equation}
where $y(\vecabstime) \in \mathbb{R}^{G}$ denotes a vector of a measured physiological quantity with measurement times $\vecabstime \in \mathbb{R}^{G}$. The baseline vector is $f^{b}(\vecabstime) \in \mathbb{R}^{G} $. There are $M$ treatments happening during the observed time period and we assume that their effects are additive. A single treatment--response vector to the $j$\textsuperscript{th} treatment is $f^{r}(\vecabstime,t_{j},\mathbf{m}_{j}) \in \mathbb{R}^{G} $, where $\mathbf{m}_{j}$ is the vector of dosages and $t_{j}$ is the time of the $j$th treatment, respectively. The noise $\boldsymbol{\epsilon}$ denotes the error term. The model in \cref{eq:model} does not differentiate between $Q$ components of a composite treatment $j$, thus we extend it in the following form to explicitly capture different responses:
\begin{equation}\label{eq:modeltypes}
    y(\vecabstime)=f^{b}(\vecabstime)+\sum_{j=1}^{M} g_{Q}(\{f_q^{r}(\vecabstime,t_{j},m_{jq})\}_{q=1}^Q)+ \boldsymbol{\epsilon},
\end{equation}
where $g_{Q}(\cdot)$ denotes a function that captures the coupling of $Q$ different responses $f_{q}^{r}(\vecabstime,t_{j},m_{jq})$ for the different components $q=1,\ldots,Q$ of the composite treatment $j$. The components of the composite treatment $j$ all share the treatment time $t_{j}$, but have varying dosages $m_{jq}$. The developed models differ in the way the $g_{Q}(\cdot)$ function treats various treatment components. Note that instead of absolute outcome times $\vecabstime$ and treatment times $t_{j}$, we define the response functions $f_q^r$ using relative times, i.e, the time difference since treatment $j$: $\vecreltime_j=\vecabstime - t_{j}$.


\subsection{Nonparametric Separate Response Model (\modelGPresp)}\label{sec:npseptt}

\begin{figure}[h]
\floatconts
  {fig:trc_types}
  {\caption{(\textbf{a}): Illustration of the treatment response in the additive model. (\textbf{b}): $\mathbf{K}^\text{TLSE}$ for the additive model. Only data in the interval $0 \leq \reltime_{jk} \leq T$ is modeled. $\mathbf{K}^\text{TLSE}$ shows the covariance between the modeled observations (blue crosses in the left plot), where the off-diagonal blocks show covariances between different treatment periods.}}
  {%
    \subfigure[Additive model][t]{\label{fig:additive}%
      \includegraphics[width=1.0\linewidth]{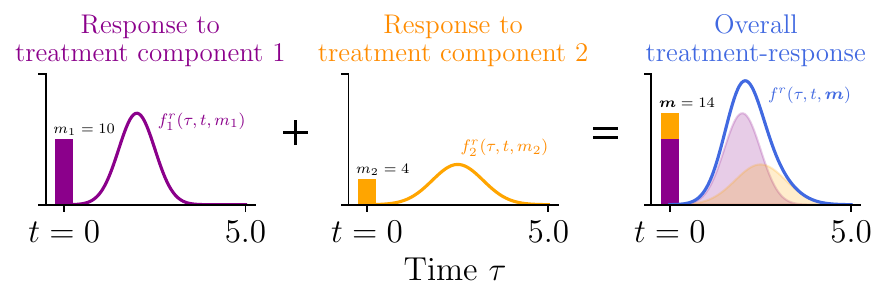}}

      \vspace{1em}
    \subfigure[$\mathbf{K}^\text{TLSE}$ illustration][b]{\label{fig:tlse}%
      \includegraphics[width=1.1\linewidth]{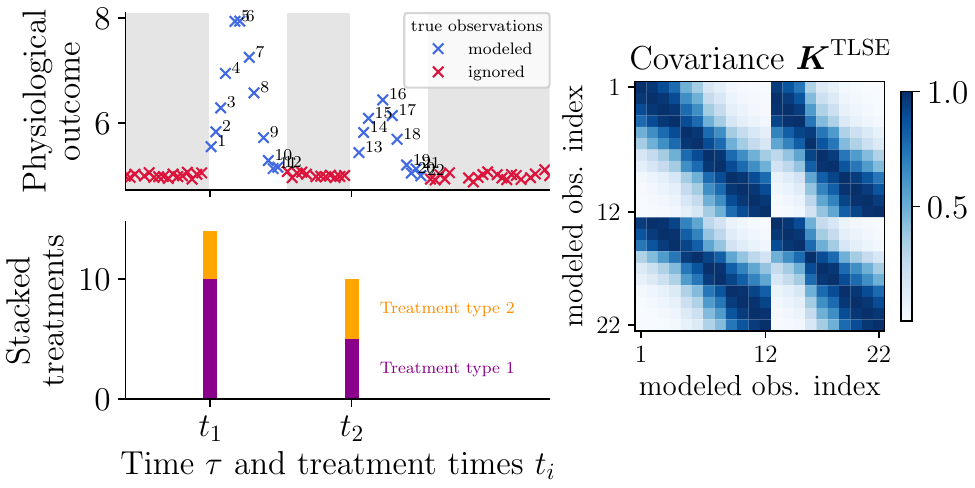}}
  }
\end{figure}

Here we describe an additive nonparametric Bayesian model, \modelGPresp, where the two primary components in \cref{eq:modeltypes}, namely the baseline and treatment--response functions, are modeled with GPs. In \modelGPresp the $g_Q(\cdot)$ is an additive function of $Q$ treatment components, resulting in
\begin{align}
\label{eq:npseptt}
      f^{r}(\vecabstime,t_{j},\textbf{m}_{j}) &= \sum_{q=1}^{Q}f_{q}^{r}(\vecabstime,t_{j},m_{jq}) \nonumber \\&= \sum_{q=1}^{Q} f_{q}^t(\vecabstime,t_{j})f_{q}^{m}(m_{jq}). 
\end{align}
The response to component $q$ of treatment $j$, $f_{q}^{r}(\vecabstime,t_{j},m_{jq})$, factorizes into $f_{q}^t(\vecabstime,t_{j})$, defining the shape of the response, and $f_{q}^{m}(m_{jq})$, denoting the magnitude (i.e., height) of the response. These functions are specific for each component $q$ of the composite treatment. \figureref{fig:trc_types} \subfigref{fig:additive} shows the sum of the responses for two treatment components ($Q=2$).

\paragraph{Covariance modeling.}
The performance and flexibility of a GP model significantly depends on its covariance function \citep{gpsrasmussen, gps_view}. One of our contributions lies in developing a covariance function that accommodates multiple individuals, while also integrating treatment dosages and time restrictions on treatment influence into the model \emph{simultaneously}. This approach enhances the model's ability to capture important TRC dependencies.

The total response function comes from the GP $y(\cdot) \sim \mathcal{GP}(0,k(\cdot,\cdot'))$, which in the case of finite input data simplifies to $\mathbf{y} = \mathcal{N}(\mathbf{0},\mathbf{K})$. The total response is the sum of the baseline function and individual treatment-response components, all having $\mathcal{GP}$ priors. Thus, the total covariance is
\begin{equation}
\label{eq:totalcov}
\mathbf{K}=\mathbf{K}^{b}+\sum_{q=1}^Q\mathbf{K}_{q}^r.
\end{equation}
The detailed derivation of $\mathbf{K}$ and its usage in training and testing is described in \appendixref{apd:third1}. 

$\mathbf{K}_{q}^r$ is structured using multi-output GPs (MOGPs) and coregionalization techniques which help in multi-task learning problems \citep{bonilla2007multi,liu2022scalable,mogpsalvarez,mogpswilk}. In our case, MOGPs allow us to work in parallel with multiple patients and treatments, sharing the information between them. The decomposition of $\mathbf{K}_{q}^r$ is derived in \appendixref{apd:third2}, leading to
\begin{equation}
\label{eq:covtrdec}
    \mathbf{K}_{q}^r=\textbf{C}_{q} \otimes \textbf{K}_{q}^\text{TLSE},
\end{equation} where $\textbf{C}_{q}$ is the coregionalization matrix, whose values are linear functions of treatment dosages, and $\textbf{K}_{q}^\text{TLSE}$ captures how different time points are correlated for a single treatment component $q$.

The Time-Limited Squared Exponential (TLSE) kernel is used to restrict the treatment influence time for the $k$\textsuperscript{th} observation, i.e.~$0 \leq \reltime_{jk} \leq T$ \citep{wright2011understanding}. This means that the time-dependent function $f_{q}^t(\abstime_k,t_{j}) \sim \mathcal{GP}(0,k_\text{TLSE}(\reltime_{jk},\reltime_{jk}'))$ with
\begin{multline}
\label{eq:tlse}
    k_\text{TLSE}(\reltime_{jk},\reltime_{jk}') = k_\text{SE}(\reltime_{jk}, \reltime_{jk}') \\
    C(\reltime_{jk}) C(\reltime_{jk}') ,
\end{multline}
where $k_\text{SE}(\reltime, \reltime') = \exp(-\frac12 (\reltime - \reltime')^2/\ell^2)$ and
\begin{equation}
    C(\reltime) = 
    \begin{cases}
      1, & \text{if } 0<\reltime<T \\
      0, & \text{otherwise}.
    \end{cases}
\end{equation}
A toy example of the observed data and the corresponding $\mathbf{K}^\text{TLSE}$ covariance matrix is shown in \figureref{fig:trc_types}\subfigref{fig:tlse}. We see that the covariance values between the closest observations are the largest. The covariance matrix captures also dependencies between observations from different treatment periods (e.g., two separate sub-sequences of consecutive modeled observations in \figureref{fig:trc_types}\subfigref{fig:tlse}). These dependencies are encoded by the non-diagonal covariance blocks.

\subsection{Nonparametric Latent Force Model (\modelGPlfm)}\label{sec:nplfm}
Incorporating the dynamics of the system into a purely data-driven model can be valuable when modeling biological time series. One approach to achieve this is the Latent force model \citep{lfms_alvarez09a}, which models how the outcome response changes with time using a linear ordinary differential equation (ODE) and then solves the ODE system \citep{edwards2000differential,boyce2021elementary} to obtain the overall response. Compared to the \modelGPresp model, this corresponds to giving a physical meaning to the response function. In \sectionref{sec:npseptt} we modeled $f_{q}^t(\abstime_k,t_{j}) \sim \mathcal{GP}(0,k_\text{TLSE}(\reltime_{jk},\reltime_{jk}'))$ directly, but here we define its differential function $(f_{q}^{t})'(\cdot,\cdot)$
\begin{equation}
\label{eq:lfm}
        (f_{q}^{t})'(\abstime_k,t_{j})=-D_{q}f_{q}^t(\abstime_k,t_{j})+S_{q}f_{q}^{l}(\abstime_{k},t_{j})
\end{equation}
with a latent force $f_{q}^{l}(\abstime_{k},t_{j}) \sim GP(0, k_\text{TLSE}(\reltime_{jk},\reltime_{jk}'))$ representing the treatment stimulus on the outcome. Solving the first-order linear ODE, we obtain the LFM kernel
\begin{align}
\label{eq:lfmcov}
    k_\text{LFM}(\reltime_{jk},\reltime_{jk}') = L \otimes L' [k_{f_{q}^{l}f_{q}^{l'}}](\reltime_{jk},\reltime_{jk}')
    \nonumber \\ = 
    S_{q}S_{q}\exp^{-D_{q}(\reltime_{jk})-D_{q}(\reltime_{jk}')} \nonumber \\ \int_{0}^{\reltime_{jk}} e^{D_{q}u}\int_{0}^{\reltime_{jk}'} e^{D_{q}u'}k_\text{TLSE}(u,u')\mathrm{d}u'\mathrm{d}u.
\end{align} with decay and sensitivity parameters $\{D_{q},S_{q} \}_{q=1}^Q$. While $D_q$ is the decline rate of the response, $S_q$ is a coupling constant, describing the sensitivity of response function to the latent power. \par
A toy example of the observed data and corresponding $\mathbf{K}^\text{LFM}$ covariance matrix is shown in \figureref{fig:kernels} of \appendixref{apd:third3}. We see that compared to the $\mathbf{K}^\text{TLSE}$ matrix in \figureref{fig:trc_types} \subfigref{fig:tlse}, the $\mathbf{K}^\text{LFM}$ has a larger covariance between the modeled outcomes which are located further away from the treatment.

\subsection{Nonparametric Convolution Model (\modelGPconv)}\label{sec:npconv}

\begin{figure*}[h]
\floatconts
  {fig:conv}
  {\caption{\modelGPconv model. By convolving the carbohydrate response $f^l(\abstime,t)$ (\textbf{left}, {\color{violet} purple}) with the fat-based filter $g(\abstime,t,m_2)$ (\textbf{middle}, {\color{orange} orange}) we obtain the total response $f^t(\abstime,t,m_2)$ (\textbf{right}, {\color{blue} blue}). We compare three composite meals with: (i) $m_2=\SI{0}{\gram}$, (ii) $m_2=\SI{4}{\gram}$ and (iii) $m_2=\SI{8}{\gram}$.}}
  {\includegraphics[width=0.8\linewidth, trim = {0 0.3cm 0 0}, clip]{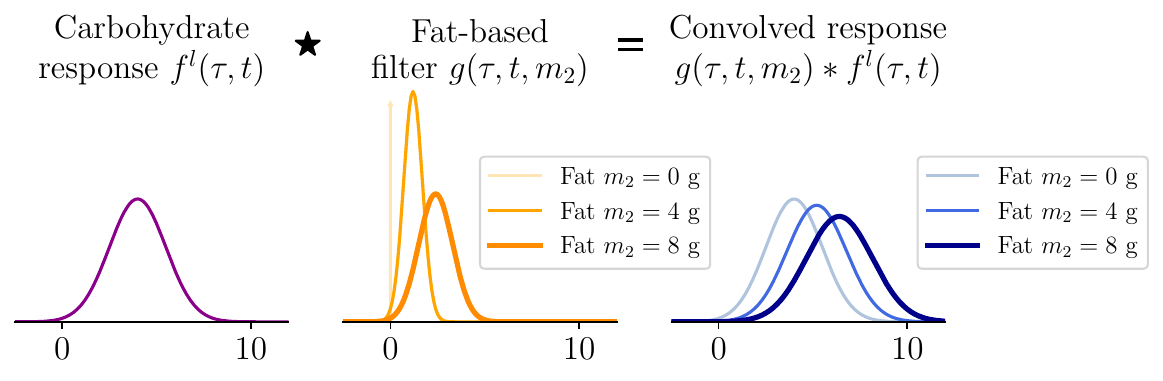}}
\end{figure*}

\modelGPresp (\cref{sec:npseptt}) and \modelGPlfm (\cref{sec:nplfm}) extend existing methods to separately model different treatment components, manage their dosages, and facilitate information sharing across patients. However, these models still combine component-specific responses additively, and hence cannot incorporate complex interactions between the components. Instead, in a more biologically motivated model the response would be mainly determined by the driving treatment component (e.g. carbohydrates), and other minor components (e.g. fat) would modify this response, for example by delaying the effect or changing its magnitude.

\paragraph{Model formulation.} One way to model the dependence between treatment components is through filtering or convolution \citep{higdon2002space,boyle2004dependent, alvarez2010efficient} operation. Here, convolution leads to an interaction of the responses to sugar and fat by taking a smoothed average of nearby time points using a smoothing filter to mix the effects of these two nutrients.

We model response as
\begin{align}
\label{eq:npconv}
      f^{r}(\abstime_k,t_{j},\textbf{m}_{j}) = f^{m}(\textbf{m}_{j}) f^{t}(\reltime_{jk}, \{m_{jq}\}_{q=2}^Q) = \nonumber \\
      \underbrace{f^{m}(\textbf{m}_{j})}_\text{Magnitude term} \int \underbrace{g(\reltime_{jk}-u, \{m_{jq}\}_{q=2}^Q)}_{\substack{\text{Smoothing} \\ \text{filter}}} \underbrace{f^{l}(u)}_{\substack{\text{Latent} \\ \text{function}}} \mathrm{d}u. 
\end{align}
The function $f^{m}(\textbf{m}_{j})$ takes dosages of all treatment components as input $f^{m}(\textbf{m}_{j})=\sum_{q=1}^Qb_{iq}m_{jq}$. The latent function $f^{l}(\cdot)$ produces a response based on one treatment component, and the filter $g(\cdot)$ modifies the response by incorporating information about other components, effectively changing the shape and location of the latent function. \figureref{fig:conv} illustrates how the convolved result is created from the filter and latent functions, with different added treatment dosages $m_2$. In this work, we use the simple Gaussian filter, which is interpretable, intuitively achieves the delaying effect of some treatment components, and maintains computational tractability.

\paragraph{Covariance derivation.}
Similarly to TLSE and LFM kernels in \sectionref{sec:npseptt} and \sectionref{sec:nplfm}, we present the convolution kernel for $f^t(\abstime_k,t_{j}) \sim \mathcal{GP}(0,k_\text{Conv}(\reltime_{jk},\reltime_{jk}'))$.
To compute the kernel analytically we use a Gaussian filter
\begin{equation}
\label{eq:gauss}
\resizebox{0.48\textwidth}{!}{
    $g(\reltime_{jk}-u, \{m_{jq}\}_{q=2}^Q) = \frac{1}{\sqrt{2\pi}\sigma} \exp \left( \frac{-(\reltime_{jk}-u-\mu)^2}{2\sigma^2} \right)$
    }
\end{equation} as the smoothing filter $g(\reltime_{jk}-u, \{m_{jq}\}_{q=2}^Q)$
and let $f^{l}(u) \sim GP(0, k_\text{TLSE}(u,u'))$. To define the filter, we assume that one treatment component has a larger influence on the final response than the other $Q-1$ components. When some $m_{jq}$ of treatment components $q=\{2, \dots, Q\}$ are non-zero, we want to shift the latent function $f^{l}(\cdot)$ with the value of $\mu$ and change its variance by adding $\sigma^2$. These shift and variance changes follow from the application of the filter function, and thus we let its parameters depend on covariates $\{m_{jq}\}_{q=2}^Q$ linearly: $\mu = \sum_{q=2}^Q\delta_q m_{jq}$ and $\sigma = \sum_{q=2}^Q\beta_q m_{jq}$. \par
After performing derivations in \appendixref{apd:fourth}, we achieve the final covariance function
\begin{align}
\label{eq:npconvkernel1}
     k_\text{Conv}(\reltime_{jk},\reltime_{jk}')=
    \frac{l}{\sqrt{l^2+\sigma^2+\sigma'^2}} \nonumber \\
    \exp \left( \frac{-1}{2} \frac{(\reltime_{jk}-\reltime_{jk}'-\mu +\mu')^2}{(l^2+\sigma^2+\sigma'^2)} \right).
\end{align}\par
A toy example of the observed data and corresponding $\mathbf{K}^\text{Conv}$ is shown in \figureref{fig:kernels} of \appendixref{apd:third3}. Compared to the $\mathbf{K}^\text{TLSE}$ in \figureref{fig:trc_types} \subfigref{fig:tlse} and $\mathbf{K}^\text{LFM}$ in \figureref{fig:kernels} of \appendixref{apd:third3}, $\mathbf{K}^\text{Conv}$ correlations between the modeled observations are more scattered and their absolute values are smaller.

\section{Experiments}
\begin{table*}[htbp!]
\floatconts
  {tab:resultstable}%
  {\caption{Comparison of methods' prediction accuracy with the real-world test set (mean $\pm$ SE, $12$ patients).}}%
  {%
\resizebox{2.0\columnwidth}{!}{%
\begin{tabular}{lcccccccc}\toprule
& \multicolumn{3}{c}{\textbf{Parametric models}} & \multicolumn{5}{c}{\textbf{Nonparametric models}}\\
\cmidrule(lr){2-4}\cmidrule(l){5-9}
           & \citet{eiv} & \modelPresp & \modelPidr & \citet{hizli2023causal} & \citet{psem_gp_lfm_cheng} & \modelGPresp & \modelGPlfm & \modelGPconv\\
\midrule
RMSE ($\downarrow$)    & 0.70 $\pm$ 0.06 & 0.68 $\pm$ 0.07 &  0.67 $\pm$ 0.06 & 0.49 $\pm$ 0.01 & 0.48 $\pm$ 0.04 & 0.45 $\pm$ 0.03 & \textbf{0.42 $\pm$ 0.02} & 0.46 $\pm$ 0.02\\
MAE ($\downarrow$)    & 0.44 $\pm$ 0.04 & 0.45 $\pm$ 0.04  & 0.43 $\pm$ 0.04  & 0.36 $\pm$ 0.01 & 0.34 $\pm$ 0.03 & 0.32 $\pm$ 0.02 & \textbf{0.30 $\pm$ 0.02} & 0.33 $\pm$ 0.02 \\
MNLL ($\downarrow$)    & 1.04 $\pm$ 0.13 & 1.11 $\pm$ 0.14  & 1.08 $\pm$ 0.13  & 0.78 $\pm$ 0.03 & 0.77 $\pm$ 0.14 & 0.94 $\pm$ 0.37 & \textbf{0.72 $\pm$ 0.20} & 0.80 $\pm$ 0.18 \\
\bottomrule
\end{tabular}
}
}
\end{table*}
Here, we demonstrate and compare the ability of the different models to derive valuable biological insights from real-world observational data, without performing costly randomized controlled trials. The code is accessible at 
\url{https://github.com/jularina/trcmed-kit}.

\subsection{Dataset}
We use real-world meal-blood glucose measurements, collected by a portable continuous glucose monitoring device and food diaries. The data (\appendixref{apd:seventh}) have been collected for non-diabetic patients, after undergoing a bariatric (e.g., gastric bypass) surgery. We include two nutrient components in the model ($Q=2$): carbohydrates (starch+sugar) and fat. The observation period is 3 days; we use the first 2 days for training and cross-validation and the last 3$^{\text{rd}}$ day for testing (\figureref{fig:cv} of \appendixref{apd:sixth}). We consider personalized predictions for $N=12$ patients. The data were provided by the Obesity Research Unit at Helsinki University. The study was approved by the Ethics Committee of Helsinki and Uusimaa Hospital District (HUS/1706/2016) and by the Helsinki University Hospital research review board (HUS269/2017).

\subsection{Comparison of prediction accuracy}
\begin{figure*}[htbp]
\floatconts
  {fig:predictions}
  {\caption{Predicted glucose response curves for a single patient by \modelGPlfm and \modelGPconv models. (\textbf{a}): \modelGPconv and \modelGPlfm results for the one-day test data of the patient. (\textbf{b}): predicted responses for one meal with average carbohydrate and fat amounts.}}
  {%
    \subfigure[Models' predictions for a single patient.][t]{\label{fig:gpmodels}%
      \includegraphics[width=0.64\linewidth]{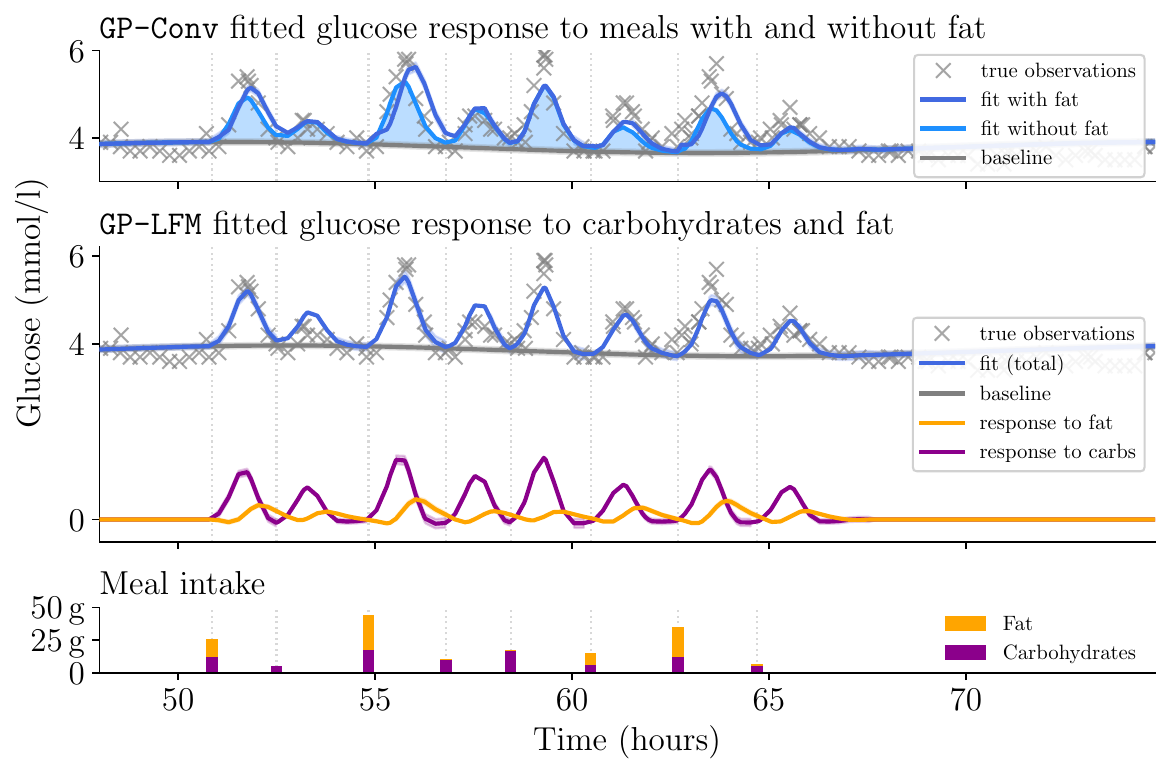}}%
    \quad
    \subfigure[Glucose response to an average meal.][b]{\label{fig:gpmodelsavg}%
      \includegraphics[width=0.33\linewidth]{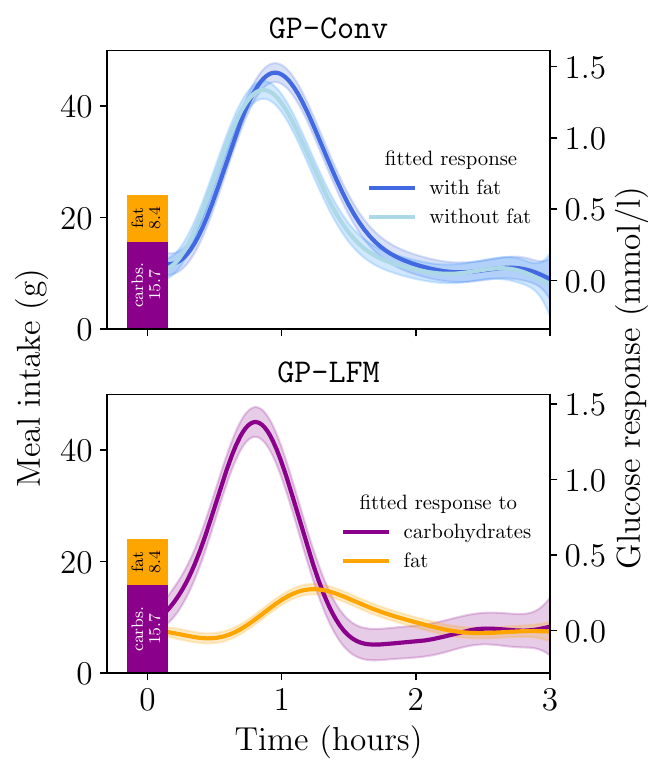}}
  }
\end{figure*}
\tableref{tab:resultstable} shows prediction accuracies on the one-day test set for all patients. We compare our models to existing approaches that are closest to the developed ones, but only incorporate one treatment component (we used carbohydrates based on domain knowledge). As metrics, we use the root mean squared error (RMSE), mean absolute error (MAE) and mean negative log-likelihood (MNLL). High standard errors for the MNLL are caused by outliers. We select a representative patient for whom we provide detailed results in \figureref{fig:predictions}, as well as \figureref{fig:latents} and \figureref{fig:predscomp} \subfigref{fig:predscomp1} of \appendixref{apd:fifth}.

Concerning the parametric approaches, we see that the existing approach by \citet{eiv} neglecting multiple treatment components performs worse than the developed \modelPresp and \modelPidr models according to RMSE metric. MAE and MNLL results are close to each other.

Overall, the nonparametric models outperform the parametric ones in all metrics. The flexibility of GPs enables modeling dependencies in the noisy and complex glucose dataset. 
We note that RMSE results for nonparametric approaches are usually within the 2 standard errors from each other. Among these, \modelGPlfm performs the best on average: it has all the advantages of the \modelGPresp, adds latent dynamics to time series modeling, and has more parameters. While \modelGPresp models the responses with GPs directly, the \modelGPlfm models latent forces for carbohydrates and fat with GPs, however, they undergo a linear transformation to create the final responses. 

The novel nonparametric convolution model \modelGPconv aims at modeling the influence of one treatment component on the response of another component, which is valuable for biological interpretability. The numerical RMSE and MAE results for it are better than for all the parametric and existing nonparametric approaches \citep{hizli2023causal, psem_gp_lfm_cheng}.
The next two subsections show detailed results for \modelGPlfm and \modelGPconv, and similar results for \modelGPresp and existing nonparametric approaches are in shown \figureref{fig:predscomp} of \appendixref{apd:fifth}.

\subsection{Detailed results for \modelGPlfm}
Here, we explore visually the best-performing \modelGPlfm model. In the middle of \figureref{fig:predictions} \subfigref{fig:gpmodels} we provide \modelGPlfm predictions for the test set of a single patient. The total glucose curve is composed of distinct responses to carbohydrates, fat, and baseline, and fits the observations accurately. To better understand the influence of carbohydrates and fat on glucose we provide response curves for a single meal with average carbohydrates and fat dosages (bottom of \figureref{fig:predictions} \subfigref{fig:gpmodelsavg}). \emph{From the biological perspective}, the results reveal that carbohydrates have a much larger and more rapid impact on glucose levels \citep{wolever1995sugars, anderson2002inverse, alsalim2016mixed}, whereas fat has a delayed and comparatively smaller influence. The model represents the mechanisms underneath the biological process of glucose generation using latent forces. 
The learnt latent force related to carbohydrate response (\figureref{fig:latents} of \appendixref{apd:fifth}) resembles the behaviour of insulin \citep{wilcox2005insulin, rahman2021role}, which is a hormone produced when blood sugar levels rise, prompting cells to absorb blood sugar. 
If the glucose peak is delayed, the insulin peak is correspondingly expected to be smaller.

\subsection{Detailed results for \modelGPconv}
Visually examining the \modelGPconv model yields useful insights. Based on the earlier findings, it is evident that fat has a relatively small influence on the glucose response. The assumption in \modelGPconv is to model the response primarily by carbohydrate intake, with a slight adjustment in the curve when fat is introduced. \emph{From the biological explainability point-of-view}, the objective is to investigate the impact of fat, compared to the prediction based solely on carbohydrates. The top of \figureref{fig:predictions} \subfigref{fig:gpmodels} shows the predicted response curve for the test data, showing (i) the fitted curve without fat, and (ii) the final curve incorporating information about all treatment components. Furthermore, \figureref{fig:predictions} \subfigref{fig:gpmodelsavg} shows \modelGPconv predictions for a  single meal with average nutrient dosages. We observe that fat in the treatment delays the response a bit and causes a higher peak later.

\section{Conclusion}
We introduced and compared several probabilistic models for TRCs with the primary focus on nonparametric models for blood glucose. The models yielded accurate personalized glucose predictions with associated uncertainties, and improved previous methods by including multiple treatment components, incorporating treatment dosages, utilizing the TLSE kernel to account for dependencies between treatments, and employing a multi-output GP to share information across patients. The nonparametric \modelGPresp and \modelGPlfm methods learned separate responses to different treatment components and combined them additively, while the novel \modelGPconv provided a biologically motivated way to understand the influence of one treatment component on the response produced by another. This model learned the physiological response primarily based on the dominant treatment component, while adjusting the response curve if another treatment component was present. In the context of blood-glucose modeling, the results revealed that carbohydrates had a significant, pronounced and rapid impact on glucose levels, whereas added fat delayed the response.

\paragraph{Limitations.}
There exist certain limitations in fitting GPs to observational data for the purpose of treatment effect estimation, in the sense of causality. For causal modeling, several assumptions, such as the no unobserved confounding (NUC), should hold. In practice, we do not expect these to hold exactly in our dataset, and therefore causal interpretations should be taken with caution. Validating the assumptions through the use of domain knowledge and with relevant sensitivity analyses \citep{robins2000sensitivity} is a valuable step toward a causal model which may be more robust and generalizable in the real-world. Another limitation is the scalability of the methods with respect to the number of inputs. Currently, the GPs scale $O(n^3)$, where $n$ is the number of observations. However, various modifications, like sparse GPs, could be used to improve the computational efficiency of the models if needed.

\paragraph{Future directions.}
One potential avenue for future research is the inclusion of additional treatment components, e.g., proteins, in the model, along with a more advanced approach to modeling interconnections between these components using the convolution technique. The convolution filter parameters may not depend linearly on the treatment dosages, but instead on some nonlinear combination, although implementing this may affect interpretability, which is one strength of the current model. Also, one more direction is connected with modeling jointly multiple response variables, for instance, with the help of MOGPs \citep{psem_gp_lfm_cheng}.

\bibliography{odnoblyudova23}

\begin{thebibliography}{43}
\providecommand{\natexlab}[1]{#1}
\providecommand{\url}[1]{\texttt{#1}}
\expandafter\ifx\csname urlstyle\endcsname\relax
  \providecommand{\doi}[1]{doi: #1}\else
  \providecommand{\doi}{doi: \begingroup \urlstyle{rm}\Url}\fi

\bibitem[Albers et~al.(2017)Albers, Levine, Gluckman, Ginsberg, Hripcsak, and
  Mamykina]{albers2017personalized}
David~J Albers, Matthew Levine, Bruce Gluckman, Henry Ginsberg, George
  Hripcsak, and Lena Mamykina.
\newblock Personalized glucose forecasting for type 2 diabetes using data
  assimilation.
\newblock \emph{PLoS Computational Biology}, 13\penalty0 (4):\penalty0
  e1005232, 2017.

\bibitem[Alsalim et~al.(2016)Alsalim, Tura, Pacini, Omar, Bizzotto, Mari, and
  Ahr{\'e}n]{alsalim2016mixed}
Wathik Alsalim, Andrea Tura, Giovanni Pacini, Bilal Omar, Roberto Bizzotto,
  Andrea Mari, and Bo~Ahr{\'e}n.
\newblock Mixed meal ingestion diminishes glucose excursion in comparison with
  glucose ingestion via several adaptive mechanisms in people with and without
  type 2 diabetes.
\newblock \emph{Diabetes, Obesity and Metabolism}, 18\penalty0 (1):\penalty0
  24--33, 2016.

\bibitem[{\'A}lvarez et~al.(2010){\'A}lvarez, Luengo, Titsias, and
  Lawrence]{alvarez2010efficient}
Mauricio {\'A}lvarez, David Luengo, Michalis Titsias, and Neil~D Lawrence.
\newblock Efficient multioutput gaussian processes through variational inducing
  kernels.
\newblock In \emph{Proceedings of the Thirteenth International Conference on
  Artificial Intelligence and Statistics}, pages 25--32. JMLR Workshop and
  Conference Proceedings, 2010.

\bibitem[Alvarez et~al.(2012)Alvarez, Rosasco, Lawrence, et~al.]{mogpsalvarez}
Mauricio~A Alvarez, Lorenzo Rosasco, Neil~D Lawrence, et~al.
\newblock Kernels for vector-valued functions: A review.
\newblock \emph{Foundations and Trends in Machine Learning}, 4\penalty0
  (3):\penalty0 195--266, 2012.

\bibitem[Anderson et~al.(2002)Anderson, Catherine, Woodend, and
  Wolever]{anderson2002inverse}
G~Harvey Anderson, Nicole~LA Catherine, Dianne~M Woodend, and Thomas~MS
  Wolever.
\newblock Inverse association between the effect of carbohydrates on blood
  glucose and subsequent short-term food intake in young men.
\newblock \emph{The American Journal of Clinical Nutrition}, 76\penalty0
  (5):\penalty0 1023--1030, 2002.

\bibitem[Balakrishnan et~al.(2014)Balakrishnan, Samavedham, and
  Rangaiah]{balakrishnan2014personalized}
Naviyn~Prabhu Balakrishnan, Lakshminarayanan Samavedham, and Gade~Pandu
  Rangaiah.
\newblock Personalized mechanistic models for exercise, meal and insulin
  interventions in children and adolescents with type 1 diabetes.
\newblock \emph{Journal of Theoretical Biology}, 357:\penalty0 62--73, 2014.

\bibitem[Bang and Robins(2005)]{bang2005doubly}
Heejung Bang and James~M Robins.
\newblock Doubly robust estimation in missing data and causal inference models.
\newblock \emph{Biometrics}, 61\penalty0 (4):\penalty0 962--973, 2005.

\bibitem[Bonilla et~al.(2007)Bonilla, Chai, and Williams]{bonilla2007multi}
Edwin~V Bonilla, Kian Chai, and Christopher Williams.
\newblock Multi-task gaussian process prediction.
\newblock \emph{Advances in Neural Information Processing Systems}, 20, 2007.

\bibitem[Boyce et~al.(2021)Boyce, DiPrima, and Meade]{boyce2021elementary}
William~E Boyce, Richard~C DiPrima, and Douglas~B Meade.
\newblock \emph{Elementary differential equations and boundary value problems}.
\newblock John Wiley \& Sons, 2021.

\bibitem[Boyle and Frean(2004)]{boyle2004dependent}
Phillip Boyle and Marcus Frean.
\newblock Dependent gaussian processes.
\newblock \emph{Advances in Neural Information Processing Systems}, 17, 2004.

\bibitem[Chen et~al.(2021)Chen, Joshi, Ghassemi, and
  Ranganath]{pml_healthcare_book}
Irene~Y. Chen, Shalmali Joshi, Marzyeh Ghassemi, and Rajesh Ranganath.
\newblock Probabilistic machine learning for healthcare.
\newblock \emph{Annual Review of Biomedical Data Science}, 4\penalty0
  (1):\penalty0 393--415, 2021.
\newblock \doi{10.1146/annurev-biodatasci-092820-033938}.
\newblock URL \url{https://doi.org/10.1146/annurev-biodatasci-092820-033938}.
\newblock PMID: 34465179.

\bibitem[Cheng et~al.(2020)Cheng, Dumitrascu, Zhang, Chivers, Draugelis, Li,
  and Engelhardt]{psem_gp_lfm_cheng}
Li-Fang Cheng, Bianca Dumitrascu, Michael Zhang, Corey Chivers, Michael
  Draugelis, Kai Li, and Barbara Engelhardt.
\newblock Patient-specific effects of medication using latent force models with
  gaussian processes.
\newblock In Silvia Chiappa and Roberto Calandra, editors, \emph{Proceedings of
  the Twenty Third International Conference on Artificial Intelligence and
  Statistics}, volume 108 of \emph{Proceedings of Machine Learning Research},
  pages 4045--4055. PMLR, 26--28 Aug 2020.
\newblock URL \url{https://proceedings.mlr.press/v108/cheng20c.html}.

\bibitem[Edwards and Penney(2000)]{edwards2000differential}
Charles~Henry Edwards and David~E Penney.
\newblock \emph{Differential equations and boundary value problems: computing
  and modeling}.
\newblock Pearson Educaci{\'o}n, 2000.

\bibitem[Ferguson(1973)]{ferguson1973bayesian}
Thomas~S Ferguson.
\newblock A bayesian analysis of some nonparametric problems.
\newblock \emph{The Annals of Statistics}, pages 209--230, 1973.

\bibitem[Gelman et~al.(2013)Gelman, Carlin, Stern, Dunson, Vehtari, and
  Rubin]{gelman2013bayesian}
Andrew Gelman, John~B Carlin, Hal~S Stern, David~B Dunson, Aki Vehtari, and
  Donald~B Rubin.
\newblock \emph{Bayesian data analysis}.
\newblock CRC press, 2013.

\bibitem[Gill and Robins(2001)]{gill2001causal}
Richard~D Gill and James~M Robins.
\newblock Causal inference for complex longitudinal data: the continuous case.
\newblock \emph{Annals of Statistics}, pages 1785--1811, 2001.

\bibitem[Gustafson(2003)]{gustafson2003measurement}
Paul Gustafson.
\newblock \emph{Measurement error and misclassification in statistics and
  epidemiology: impacts and Bayesian adjustments}.
\newblock CRC Press, 2003.

\bibitem[Hall(2008)]{hall2008measurement}
Daniel~B Hall.
\newblock Measurement error in nonlinear models: a modern perspective, 2008.

\bibitem[Higdon(2002)]{higdon2002space}
Dave Higdon.
\newblock Space and space-time modeling using process convolutions.
\newblock In \emph{Quantitative methods for current environmental issues},
  pages 37--56. Springer, 2002.

\bibitem[H{\i}zl{\i} et~al.(2023)H{\i}zl{\i}, John, Juuti, Saarinen,
  Pietil{\"a}inen, and Marttinen]{hizli2023causal}
{\c{C}}a{\u{g}}lar H{\i}zl{\i}, ST~John, Anne~Tuulikki Juuti, Tuure~Tapani
  Saarinen, Kirsi~Hannele Pietil{\"a}inen, and Pekka Marttinen.
\newblock Causal modeling of policy interventions from treatment-outcome
  sequences.
\newblock In \emph{International Conference on Machine Learning}, pages
  13050--13084. PMLR, 2023.

\bibitem[Lawrence et~al.(2006)Lawrence, Sanguinetti, and
  Rattray]{gps_transcriptionalreg}
Neil Lawrence, Guido Sanguinetti, and Magnus Rattray.
\newblock Modelling transcriptional regulation using gaussian processes.
\newblock In B.~Sch\"{o}lkopf, J.~Platt, and T.~Hoffman, editors,
  \emph{Advances in Neural Information Processing Systems}, volume~19. MIT
  Press, 2006.
\newblock URL
  \url{https://proceedings.neurips.cc/paper/2006/file/f42c7f9c8aeab0fc412031e192e2119d-Paper.pdf}.

\bibitem[Liu et~al.(2022)Liu, Ding, Xie, Jiang, Zhao, and
  Wang]{liu2022scalable}
Haitao Liu, Jiaqi Ding, Xinyu Xie, Xiaomo Jiang, Yusong Zhao, and Xiaofang
  Wang.
\newblock Scalable multi-task gaussian processes with neural embedding of
  coregionalization.
\newblock \emph{Knowledge-Based Systems}, 247:\penalty0 108775, 2022.

\bibitem[Lunceford et~al.(2002)Lunceford, Davidian, and
  Tsiatis]{lunceford2002estimation}
Jared~K Lunceford, Marie Davidian, and Anastasios~A Tsiatis.
\newblock Estimation of survival distributions of treatment policies in
  two-stage randomization designs in clinical trials.
\newblock \emph{Biometrics}, 58\penalty0 (1):\penalty0 48--57, 2002.

\bibitem[M{\"u}ller and Mitra(2013)]{muller2013bayesian}
Peter M{\"u}ller and Riten Mitra.
\newblock Bayesian nonparametric inference--why and how.
\newblock \emph{Bayesian Analysis (Online)}, 8\penalty0 (2), 2013.

\bibitem[Murphy et~al.(2007{\natexlab{a}})Murphy, Collins, and
  Rush]{murphy2007customizing}
Susan~A Murphy, Linda~M Collins, and A~John Rush.
\newblock Customizing treatment to the patient: Adaptive treatment strategies.
\newblock \emph{Drug and Alcohol Dependence}, 88\penalty0 (Suppl 2):\penalty0
  S1, 2007{\natexlab{a}}.

\bibitem[Murphy et~al.(2007{\natexlab{b}})Murphy, Lynch, Oslin, McKay, and
  TenHave]{murphy2007developing}
Susan~A Murphy, Kevin~G Lynch, David Oslin, James~R McKay, and Tom TenHave.
\newblock Developing adaptive treatment strategies in substance abuse research.
\newblock \emph{Drug and Alcohol Dependence}, 88:\penalty0 S24--S30,
  2007{\natexlab{b}}.

\bibitem[Rahman et~al.(2021)Rahman, Hossain, Das, Kundu, Adegoke, Rahman,
  Hannan, Uddin, and Pang]{rahman2021role}
Md~Saidur Rahman, Khandkar~Shaharina Hossain, Sharnali Das, Sushmita Kundu,
  Elikanah~Olusayo Adegoke, Md~Ataur Rahman, Md~Abdul Hannan, Md~Jamal Uddin,
  and Myung-Geol Pang.
\newblock Role of insulin in health and disease: an update.
\newblock \emph{International Journal of Molecular Sciences}, 22\penalty0
  (12):\penalty0 6403, 2021.

\bibitem[Rasmussen and Williams(2005)]{gpsrasmussen}
Carl~Edward Rasmussen and Christopher K.~I. Williams.
\newblock \emph{{Gaussian Processes for Machine Learning}}.
\newblock MIT Press, 11 2005.
\newblock ISBN 9780262256834.
\newblock \doi{10.7551/mitpress/3206.001.0001}.
\newblock URL \url{https://doi.org/10.7551/mitpress/3206.001.0001}.

\bibitem[Robins(1987)]{robins1987graphical}
James Robins.
\newblock A graphical approach to the identification and estimation of causal
  parameters in mortality studies with sustained exposure periods.
\newblock \emph{Journal of Chronic Diseases}, 40:\penalty0 139S--161S, 1987.

\bibitem[Robins et~al.(2000{\natexlab{a}})Robins, Hernan, and
  Brumback]{robins2000marginal}
James~M Robins, Miguel~Angel Hernan, and Babette Brumback.
\newblock Marginal structural models and causal inference in epidemiology.
\newblock \emph{Epidemiology}, pages 550--560, 2000{\natexlab{a}}.

\bibitem[Robins et~al.(2000{\natexlab{b}})Robins, Rotnitzky, and
  Scharfstein]{robins2000sensitivity}
James~M Robins, Andrea Rotnitzky, and Daniel~O Scharfstein.
\newblock Sensitivity analysis for selection bias and unmeasured confounding in
  missing data and causal inference models.
\newblock \emph{IMA Volumes In Mathematics and Its Applications}, 116:\penalty0
  1--94, 2000{\natexlab{b}}.

\bibitem[Rodriguez and M{\"u}ller(2013)]{rodriguez2013nonparametric}
Abel Rodriguez and Peter M{\"u}ller.
\newblock Nonparametric bayesian inference.
\newblock In \emph{NSF-CBMS Regional Conference Series in Probability and
  Statistics}, volume~9, pages i--110. JSTOR, 2013.

\bibitem[Scholkopf and Smola(2018)]{gps_view}
Bernhard Scholkopf and Alexander~J Smola.
\newblock \emph{Learning with kernels: support vector machines, regularization,
  optimization, and beyond}.
\newblock MIT press, 2018.

\bibitem[Silva(2016)]{silva2016observational}
Ricardo Silva.
\newblock Observational-interventional priors for dose-response learning.
\newblock \emph{Advances in Neural Information Processing Systems}, 29, 2016.
\newblock URL
  \url{https://proceedings.neurips.cc/paper_files/paper/2016/file/aff1621254f7c1be92f64550478c56e6-Paper.pdf}.

\bibitem[Soleimani et~al.(2017)Soleimani, Subbaswamy, and
  Saria]{trm_counterfactual_reasoning}
Hossein Soleimani, Adarsh Subbaswamy, and Suchi Saria.
\newblock Treatment-response models for counterfactual reasoning with
  continuous-time, continuous-valued interventions.
\newblock In \emph{Uncertainty in Artificial Intelligence - Proceedings of the
  33rd Conference, UAI 2017}. AUAI Press Corvallis, 2017.
\newblock 33rd Conference on Uncertainty in Artificial Intelligence, UAI 2017 ;
  Conference date: 11-08-2017 Through 15-08-2017.

\bibitem[Spiegelhalter et~al.(1999)Spiegelhalter, Myles, Jones, and
  Abrams]{spiegelhalter1999introduction}
David~J Spiegelhalter, Jonathan~P Myles, David~R Jones, and Keith~R Abrams.
\newblock An introduction to bayesian methods in health technology assessment.
\newblock \emph{BMJ}, 319\penalty0 (7208):\penalty0 508--512, 1999.

\bibitem[Van~der Wilk et~al.(2020)Van~der Wilk, Dutordoir, John, Artemev, Adam,
  and Hensman]{mogpswilk}
Mark Van~der Wilk, Vincent Dutordoir, ST~John, Artem Artemev, Vincent Adam, and
  James Hensman.
\newblock A framework for interdomain and multioutput {G}aussian processes,
  2020.
\newblock URL \url{https://arxiv.org/abs/2003.01115}.

\bibitem[Wilcox(2005)]{wilcox2005insulin}
Gisela Wilcox.
\newblock Insulin and insulin resistance.
\newblock \emph{Clinical Biochemist Reviews}, 26\penalty0 (2):\penalty0 19,
  2005.

\bibitem[Wolever and Miller(1995)]{wolever1995sugars}
TM~Wolever and J~Brand Miller.
\newblock Sugars and blood glucose control.
\newblock \emph{The American Journal of Clinical Nutrition}, 62\penalty0
  (1):\penalty0 212S--221S, 1995.

\bibitem[Wright et~al.(2011)Wright, Winter, and
  Duffull]{wright2011understanding}
Daniel~FB Wright, Helen~R Winter, and Stephen~B Duffull.
\newblock Understanding the time course of pharmacological effect: a pkpd
  approach.
\newblock \emph{British Journal of Clinical Pharmacology}, 71\penalty0
  (6):\penalty0 815--823, 2011.

\bibitem[Xu et~al.(2016)Xu, Xu, and Saria]{pmlr-v56-Xu16}
Yanbo Xu, Yanxun Xu, and Suchi Saria.
\newblock A non-parametric bayesian approach for estimating treatment-response
  curves from sparse time series.
\newblock In Finale Doshi-Velez, Jim Fackler, David Kale, Byron Wallace, and
  Jenna Wiens, editors, \emph{Proceedings of the 1st Machine Learning for
  Healthcare Conference}, volume~56 of \emph{Proceedings of Machine Learning
  Research}, pages 282--300, Northeastern University, Boston, MA, USA, 18--19
  Aug 2016. PMLR.
\newblock URL \url{https://proceedings.mlr.press/v56/Xu16.html}.

\bibitem[Zhang et~al.(2020)Zhang, A.Ashrafi, Juuti, Pietilainen, and
  Marttinen]{eiv}
Guangyi Zhang, Reza A.Ashrafi, Anne Juuti, Kirsi Pietilainen, and Pekka
  Marttinen.
\newblock Errors-in-variables modeling of personalized treatment-response
  trajectories.
\newblock \emph{IEEE Journal of Biomedical and Health Informatics},
  PP:\penalty0 1--1, 04 2020.
\newblock \doi{10.1109/JBHI.2020.2987323}.

\bibitem[Álvarez et~al.(2009)Álvarez, Luengo, and Lawrence]{lfms_alvarez09a}
Mauricio Álvarez, David Luengo, and Neil~D. Lawrence.
\newblock Latent force models.
\newblock In David van Dyk and Max Welling, editors, \emph{Proceedings of the
  Twelth International Conference on Artificial Intelligence and Statistics},
  volume~5 of \emph{Proceedings of Machine Learning Research}, pages 9--16,
  Hilton Clearwater Beach Resort, Clearwater Beach, Florida USA, 16--18 Apr
  2009. PMLR.
\newblock URL \url{https://proceedings.mlr.press/v5/alvarez09a.html}.

\end{thebibliography}

\appendix

\section{Parametric models}\label{apd:second}
To generate baseline scores for the more advanced nonparametric models, we initially study parametric Bayesian models. To understand the influence of treatments on the physiological state, we model the treatment--response curve for patient $i$ using a bell-shaped parametric function \citep{eiv}.
\begin{equation}
\label{eq:bellcurve}
    f^{r}(\vecabstime,t_{j},\mathbf{m}_{j}) = h_{ij} \exp\left(\frac{-0.5(\vecabstime-t_{j}-3l_i)^2}{l_{i}^2}\right)
\end{equation}
This function has two parameters, $h_{ij}$ controls the height of the curve (magnitude of the response) and $l_{i}$ controls the width (reflecting the duration of the response).

The hierarchical Bayesian models \citep{gelman2013bayesian} are used, imposing the connection between the individuals by specifying the same hyperparameter for sampling height and width parameters, but at the same time treating these coefficients individually-specific.

We aim to distinguish between different treatment components in this paper and to achieve this goal, we develop \modelPresp and \modelPidr parametric models that incorporate distinct connections among responses to disparate treatment categories.
\subsection*{Separate response model (\modelPresp)}
In this scenario, multiple response functions are considered, with each one corresponding to different treatment components $Q$. The overall response function is decomposed in the following manner
\begin{align}
\label{eq:pshr}
    f^{r}(\vecabstime,t_{j},\textbf{m}_{j}) = \sum_{q=1}^Q f_{q}^{r}(\vecabstime,t_{j},m_{jq}) = \nonumber \\ \sum_{q=1}^Q h_{ijq} \exp\left(\frac{-0.5(\vecabstime-t_{j}-3l_{iq})^2}{l_{iq}^2}\right)
\end{align} with $h_{ijq}=\beta_{iq} m_{jq}$. There are curve parameters, corresponding to each component of treatment. It helps to catch various differences in the ways the specific treatment component influences the overall physiological quantity.
\subsection*{Individualized dependent response model (\modelPidr)}
In this model, the lengthscale parameter $l$, which determines the width of the curves, is dependent between different treatment components. Specifically, the widths of the $Q-1$ treatment components are dependent on the width of the first treatment component, with unique coefficients $c_{q}$ for each $q$. Considering the aforementioned dependencies, we derive the following equation
\begin{align}
\label{eq:pidr}
    f^{r}(\vecabstime,t_{j},\textbf{m}_{j}) = \sum_{q=1}^Q f_{q}^{r}(\vecabstime,t_{j},m_{jq}) =
    \nonumber \\ 
    h_{ij1} \exp\left(\frac{-0.5(\vecabstime-t_{j}-3l_{i1})^2}{l_{i1}^2}\right) + \nonumber \\ \sum_{q=2}^{Q} h_{ijq} \exp\left(\frac{-0.5(\vecabstime-t_{j}-3c_{q}l_{i1})^2}{(c_{q}l_{i1})^2}\right)
\end{align} with $h_{ijq}=\beta_{iq} m_{jq}$. We account for the possible dependence of the width of the bell-shaped response functions for various treatment components, which in \modelPresp used to be completely independent.

\section{Covariance for nonparametric models}\label{apd:third}
\subsection{Inferring the posteriors for the nonparametric models}\label{apd:third1}
In the finite input data case $\mathbf{y} = \mathbf{f}^{b}+\sum_{q=1}^Q \mathbf{f}_{q}^r + \boldsymbol{\epsilon}$ with $\mathbf{f}^{b} = \mathcal{N}(\mathbf{0},\mathbf{K}^{b}$), $\mathbf{f}_{q}^r = \mathcal{N}(\mathbf{0},\mathbf{K}_{q}^r$) and $\mathbf{y} = \mathcal{N}(\mathbf{0},\mathbf{K})$. Our goal is to work with all-patient, all-treatment data at once, meaning that if the input data contains $P$ patients with $G_i$ observations for each patient $i$, then $\mathbf{y} \in R^s$ with $s=\sum_{i=1}^P G_i$. The full covariance is
\begin{align}
\label{eq:totalcovderiv}
    \mathbf{K}=\cov(\mathbf{y},\mathbf{y'}) = E[(\mathbf{y}-E(\mathbf{y}))(\mathbf{y'}-E(\mathbf{y'}))] \nonumber \\ =E[(\mathbf{f}^{b}+\sum_{q=1}^Q\mathbf{f}_{q}^r)(\mathbf{f}^{b'}+\sum_{q=1}^Q\mathbf{f}_{q}^{r'})] \nonumber \\ =\cov(\mathbf{f}^{b},\mathbf{f}^{b'})+\sum_{q=1}^Q\cov(\mathbf{f}_{q}^r,\mathbf{f}_{q}^{r'})=\mathbf{K}^{b}+\sum_{q=1}^Q\mathbf{K}_{q}^r.
\end{align}

To make predictions for the test data $X^{*}$, we infer the posteriors for the baseline and treatment--response functions separately in order to understand the individual influence of each part on the total curve prediction. The equations for train and test parts are
\begin{align}
\label{eq:nonparamgpstrain}
    \mathbf{y} &= \mathbf{f}^{b}+\sum_{q=1}^Q\mathbf{f}_{q}^{r}+\boldsymbol{\epsilon}, \\
    \mathbf{f}^{\ast} &= \mathbf{f}^{b\ast}+\sum_{q=1}^Q\mathbf{f}_{q}^{r\ast},
\end{align}
respectively. The goal is to sample $\mathbf{f}_{b}^\text{Test}$ and $\mathbf{f}_{rq}^\text{Test}$ from the GP posterior. For the baseline, we consider the joint distribution
\begin{equation}
\label{eq:nonparamgpsinferbase}
	\begin{bmatrix} 
		\mathbf{y} \\ 
		\mathbf{f}^{b\ast} \\ 
	\end{bmatrix} \sim 
 \mathcal{N}\left(\begin{bmatrix} 
		\mathbf{0} \\ 
		\mathbf{0} \\ 
	\end{bmatrix},
 \begin{bmatrix} 
		\mathbf{K}(X,X)+\sigma_{y}^2 \mathbf{I} & \mathbf{K}^{b}(X,X^{\ast}) \\ 
		\mathbf{K}^{b}(X^{\ast},X) & \mathbf{K}^{b}(X^{\ast},X^{\ast}) \\ 
	\end{bmatrix}\right)
\end{equation}
which gives the posterior 
\begin{align}
\label{eq:nonparamgpsinferbase1}
\mathbf{f}^{b\ast} \sim \mathcal{N}(\mathbf{\overline{f}}^{b\ast},\cov(\mathbf{f}^{b\ast},\mathbf{f}^{b\ast'})), \\
\mathbf{\overline{f}}^{b\ast} = \mathbf{K}^{b}(X^{\ast},X) (\mathbf{K}(X,X)+\sigma_{y}^2 \mathbf{I})^{-1} \mathbf{y}, \\
\cov(\mathbf{f}^{b\ast},\mathbf{f}^{b\ast'}) = \mathbf{K}^{b}(X^{\ast},X^{\ast}) - \mathbf{K}^{b}(X^{\ast},X) \nonumber \\ (\mathbf{K}(X,X)+\sigma_{y}^2 \mathbf{I})^{-1} \mathbf{K}^{b}(X,X^{\ast}).
\end{align}
For the response function of treatment component $q$ we come from
\begin{equation}
\label{eq:nonparamgpsinferr}
	\begin{bmatrix} 
		\mathbf{y} \\ 
		\mathbf{f}_{q}^{r\ast} \\ 
	\end{bmatrix} \sim 
 \mathcal{N}\left(\begin{bmatrix} 
		\mathbf{0} \\ 
		\mathbf{0} \\ 
	\end{bmatrix},
 \begin{bmatrix} 
		\mathbf{K}(X,X)+\sigma_{y}^2 \mathbf{I} & \mathbf{K}_q^{r}(X,X^{\ast}) \\ 
		\mathbf{K}_q^{r}(X^{\ast},X) & \mathbf{K}_q^{r}(X^{\ast},X^{\ast}) \\ 
	\end{bmatrix}\right)
\end{equation}
to the posterior
\begin{align}
\label{eq:nonparamgpsinferr1}
\mathbf{f}_{q}^{r\ast} \sim N(\mathbf{\overline{f}}_{q}^{r\ast},\cov(\mathbf{f}_{q}^{r\ast},\mathbf{f}_{q}^{r\ast'})), \\
\mathbf{\overline{f}}_{q}^{r\ast} = \mathbf{K}_{q}^r(X^{*},X) (\mathbf{K}(X,X)+\sigma_{y}^2 \mathbf{I})^{-1} \mathbf{y}, \\
\cov(\mathbf{f}_{q}^{r\ast},\mathbf{f}_{q}^{r\ast'}) = \mathbf{K}_{q}^r(X^{\ast},X^{\ast}) - \mathbf{K}_{q}^r(X^{\ast},X) \nonumber \\ (\mathbf{K}(X,X)+\sigma_{y}^2 \mathbf{I})^{-1} \mathbf{K}_{q}^r(X,X^{\ast}).
\end{align}

\subsection{Multi-output GPs approach for covariance construction}\label{apd:third2}
It is important to understand how the covariance for the response $\mathbf{f}_{q}^{r}$ of the concrete treatment component $q$ is structured. Here $\mathbf{f}_{q}^{r}=\mathbf{f}_{q}^{m}\mathbf{f}_{q}^{t}=\textbf{W}_{q}\mathbf{f}_{q}^{t}$, where $\textbf{W}_{q}=\textbf{b}_{q}*\textbf{m}_{q}$ depends on the treatment dosages and $\mathbf{f}_{q}^{t}$ depends on the differences between observation and treatment times. The vectorized parameter $\textbf{b}_{q}$ includes the individual-specific parameters $\{b_{iq}\}_{i=1}^P$. To calculate $\mathbf{K}_{q}^r$, we use MOGPs and the linear model of coregionalization approach
\begin{equation}
    \mathbf{K}_{q}^r = \cov(\textbf{W}_{q}\mathbf{f}_{q}^{t},\textbf{W}_{q}'\mathbf{f}_{q}^{t'}) = \textbf{C}_q \otimes \mathbf{K}_{q}^\text{TLSE} ,
\end{equation}
where $\textbf{C}_q=\textbf{W}_q\textbf{W}_q^{'T}+\diag(k)$.
By modeling all the patients and treatments at once, we can include the correlations between them in the model to make the process more informed.

\subsection{Covariance matrices for artificial dataset}\label{apd:third3}
The visualizations of the covariance matrices for the models \modelGPlfm and \modelGPconv on the toy data are provided in \figureref{fig:kernels}.
\begin{figure}[h]
\floatconts
  {fig:kernels}
  {\caption{$\mathbf{K}^\text{LFM}$ and $\mathbf{K}^\text{Conv}$ creation for the suitable observed data, lying within the interval $0 \leq \reltime_{jk} \leq T$, is shown. Kernels show covariance between outcomes. On the $x$- and $y$-axes there are the indices of the modeled observations (blue crosses in the upper plot).}}
  {\includegraphics[width=1.0\linewidth]{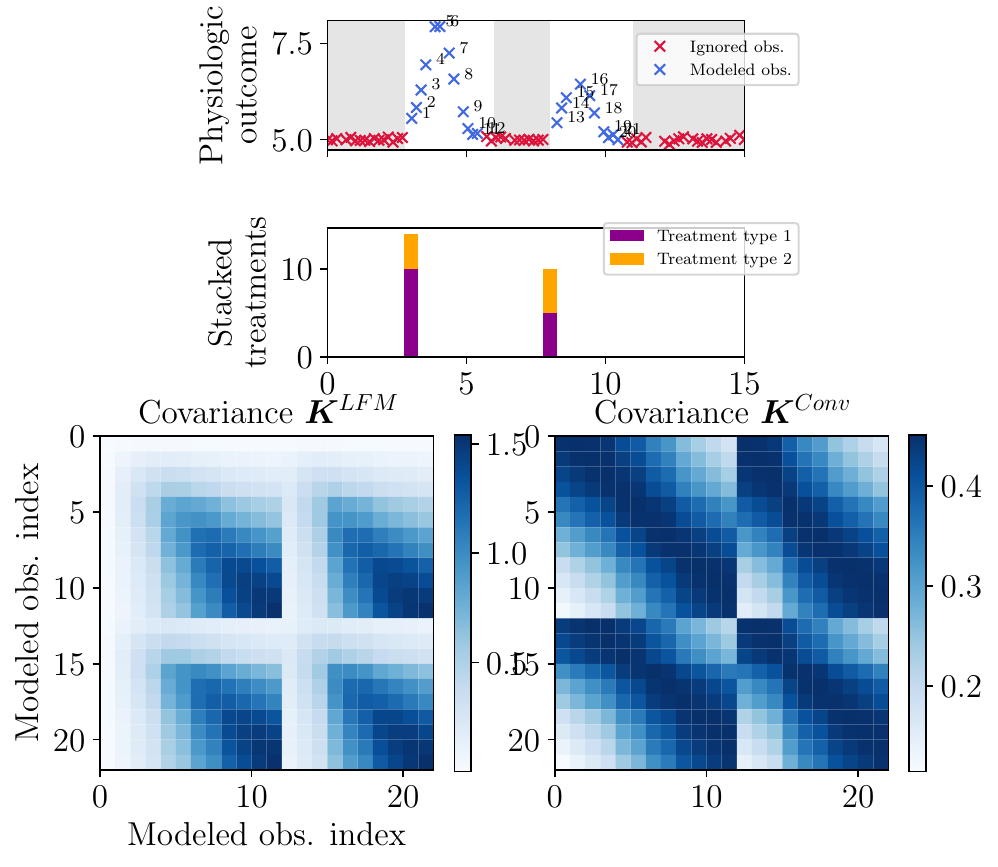}}
\end{figure}

\section{Covariance for convolution derivation}\label{apd:fourth}
We want to calculate covariance for the  
\begin{align*}
      f^{t}(\reltime_{jk}) = \int g(\reltime_{jk}-u) f^{l}(u)\mathrm{d}u. 
\end{align*}
In that case the kernel is
\begin{gather}
     k_\text{Conv}(\reltime_{jk},\reltime_{jk}') =\iint g(\reltime_{jk}-u) g'(\reltime_{jk}'-u') \nonumber \\ \cov(f^{l}(u), f^{l'}(u')) \mathrm{d}u' \mathrm{d}u = \int g(\reltime_{jk}-u) \nonumber \\ \int g'(\reltime_{jk}'-u') \cov(f^{l}(u), f^{l'}(u')) \mathrm{d}u' \mathrm{d}u
\end{gather}
with 
\begin{equation*}
    g(\reltime_{jk}-u) = \frac{1}{\sqrt{2\pi}\sigma} \exp \left(\frac{-1}{2\sigma^2}(\reltime_{jk}-u-\mu)^2 \right)
\end{equation*}
and 
\begin{equation*}
    \cov(f^{l}(u), f^{l'}(u')) = \exp \left( \frac{-1}{2} \frac{(u-u')^2}{l^2} \right).
\end{equation*}
First, we have to evaluate the integral
\begin{gather}
     \int g'(\reltime_{jk}'-u') \cov(f^{l}(u), f^{l'}(u')) \mathrm{d}u' \nonumber \\ = \int \frac{1}{\sqrt{2\pi}\sigma'} \exp \left(\frac{-1}{2\sigma'^2}(\reltime_{jk}'-u'-\mu')^2 \right) \nonumber \\ \exp \left( \frac{-1}{2} \frac{(u-u')^2}{l^2} \right) \mathrm{d}u' \nonumber = 
     \int \frac{1}{\sqrt{2\pi}\sigma'} \\ \exp \left( \frac{-1}{2} \underbrace{\left[ \sigma'^{-2}(\reltime_{jk}'-u'-\mu')^2 + l^{-2} (u-u')^2 \right]}_{\ast} \right) \mathrm{d}u' ,
\end{gather}
where the term $\ast$ is calculated by completing the square:
\begin{multline}
\sigma'^{-2}(\reltime_{jk}'-u'-\mu')^2 + l^{-2} (u-u')^2 \\ = (\sigma'^{-2} + l^{-2})(u'-a)^2 + b (u-(\reltime_{jk}'-\mu'))^2
\end{multline}
with $a=(\sigma'^{-2} + l^{-2})^{-1} (\sigma'^{-2}[\reltime_{jk}'-\mu'] + l^{-2}u)$ and $b=\sigma'^{-2}(\sigma'^{-2} + l^{-2})^{-1}l^{-2}$. Accordingly, we obtain the first integral as
\begin{multline}
\int g'(\reltime_{jk}'-u') \cov(f^{l}(u), f^{l'}(u')) \mathrm{d}u' \\ = \frac{1}{\sigma' \sqrt{\sigma'^{-2} + l^{-2}}} \exp \left( \frac{-1}{2} \left[ b (u-(\reltime_{jk}'-\mu'))^2 \right] \right).
\end{multline}
The second integral is
\begin{gather}
\frac{1}{\sigma' \sqrt{\sigma'^{-2} + l^{-2}}} \int \frac{1}{\sqrt{2\pi}\sigma} \exp \left(\frac{-1}{2\sigma^2}(\reltime_{jk}-u-\mu)^2 \right) \nonumber \\
\exp \left( \frac{-1}{2} \left[ b (u-(\reltime_{jk}'-\mu'))^2 \right] \right) \mathrm{d}u \nonumber \\ = \frac{1}{\sigma' \sqrt{\sigma'^{-2} + l^{-2}}} \int \frac{1}{\sqrt{2\pi}\sigma} \nonumber \\ \exp \left( \frac{-1}{2} \underbrace{\left[ \sigma^{-2}(\reltime_{jk}-u-\mu)^2 + b (u-(\reltime_{jk}'-\mu'))^2 \right]}_{\ast\ast} \right) \mathrm{d}u ,
\end{gather}
where for $\ast\ast$ the completion of the square is
\begin{multline}
\sigma^{-2}(\reltime_{jk}-u-\mu)^2 + b (u-(\reltime_{jk}'-\mu'))^2 \\ = (\sigma^{-2} + b)(u-c)^2 + d (\reltime_{jk}-\mu-(\reltime_{jk}'-\mu'))^2    
\end{multline}
with $c=(\sigma^{-2} + b)^{-1} (\sigma^{-2}[\reltime_{jk}-\mu] + b[\reltime_{jk}'-\mu'])$ and $d=(\sigma^{2}+\sigma'^{2}+l^2)^{-1}$. As follows, we receive the second integral and the whole kernel as
\begin{gather}
     k_\text{Conv}(\reltime_{jk},\reltime_{jk}')=
     \frac{1}{\sigma' \sqrt{(\sigma'^{-2} + l^{-2})(\sigma^{-2}+b)}} \nonumber \\
     \exp \left( \frac{-1}{2} \frac{(\reltime_{jk}-\reltime_{jk}'-\mu +\mu')^2}{(l^2+\sigma^2+\sigma'^2)} \right) = 
    \frac{l}{\sqrt{l^2+\sigma^2+\sigma'^2}} \nonumber \\
    \exp \left( \frac{-1}{2} \frac{(\reltime_{jk}-\reltime_{jk}'-\mu +\mu')^2}{(l^2+\sigma^2+\sigma'^2)} \right).
\end{gather}

\section{Original data description}\label{apd:seventh}
In \tableref{tab:statstable} we present summarized statistics of the original dataset. We also provide the visual description of the real-world data: aggregated in \figureref{fig:summary_statistics_plot} and for each concrete patient in \figureref{fig:summary_statistics_plot_indiv}.

\begin{table*}[htbp!]
\floatconts
  {tab:statstable}%
  {\caption{Original dataset statistics.}}%
  {%
\resizebox{2.0\columnwidth}{!}{%
\begin{tabular}{|r|c|c|c|c|c|c|c|c|c|c|}\hline%
 & Glucose \# train & Treatment \# train & Glucose median train& Carbs median train& Fat median train& Glucose \# test & Treatment \# test & Glucose median test& Carbs median test & Fat median test\\\hline\hline
\begin{filecontents*}{images/summary_statistics_data.csv}
glucnumtrain,treatnumtrain,glucmedtrain,carbsmedtrain,fatmedtrain,glucnumtest,treatnumtest,glucmedtest,carbsmedtest,fatmedtest
173,14,4.2,15.62,1.73,129,7,4.1,11.14,2.14
164,16,5.1,8.29,1.5,71,6,4.8,8.31,3.99
172,15,3.9,11.44,7.92,120,8,3.6,9.3,4.11
248,15,4.0,11.85,4.38,180,8,3.9,11.05,4.96
221,20,4.3,9.2,1.92,154,9,4.3,11.94,2.07
187,14,4.5,19.26,7.48,130,6,3.9,17.4,2.58
160,10,3.9,18.13,3.91,130,6,3.7,14.31,2.93
203,15,4.8,10.93,0.27,146,7,4.7,9.04,5.64
169,17,3.8,10.2,5.33,130,9,3.7,10.2,6.5
201,20,4.9,10.21,2.4,154,8,4.6,13.05,3.82
182,25,5.2,8.16,1.81,125,11,5.0,8.6,0.5
120,11,4.1,9.8,2.25,88,5,3.9,26.4,3.3
\end{filecontents*}
\csvreader[
late after line = \\\hline
]{images/summary_statistics_data.csv}%
{glucnumtrain=\glucnumtrain, treatnumtrain=\treatnumtrain, glucmedtrain=\glucmedtrain, carbsmedtrain=\carbsmedtrain, fatmedtrain=\fatmedtrain,
glucnumtest=\glucnumtest, treatnumtest=\treatnumtest, glucmedtest=\glucmedtest, carbsmedtest=\carbsmedtest, fatmedtest=\fatmedtest
}{%
\thecsvrow & \glucnumtrain & \treatnumtrain & \glucmedtrain & \carbsmedtrain & \fatmedtrain & \glucnumtest & \treatnumtest & \glucmedtest & \carbsmedtest & \fatmedtest
}%
\end{tabular}
}
}
\end{table*}

\begin{figure}[htb]
\floatconts
  {fig:summary_statistics_plot}
  {\caption{Original data distribution, aggregated for all the patients.}}
  {\includegraphics[width=1.0\linewidth, trim={0 0.15cm 0 0}, clip]{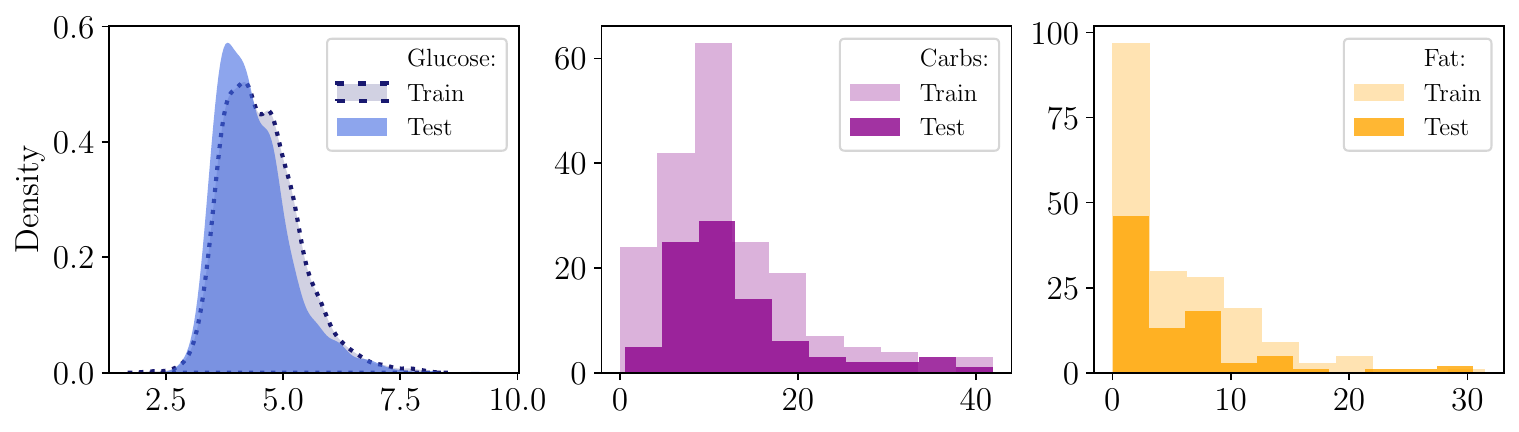}}
\end{figure}

\begin{figure}[htb]
\floatconts
  {fig:summary_statistics_plot_indiv}
  {\caption{Original data distribution for each individual patient.}}
  {\includegraphics[width=1.0\linewidth, trim={0 0.15cm 0 0}, clip]{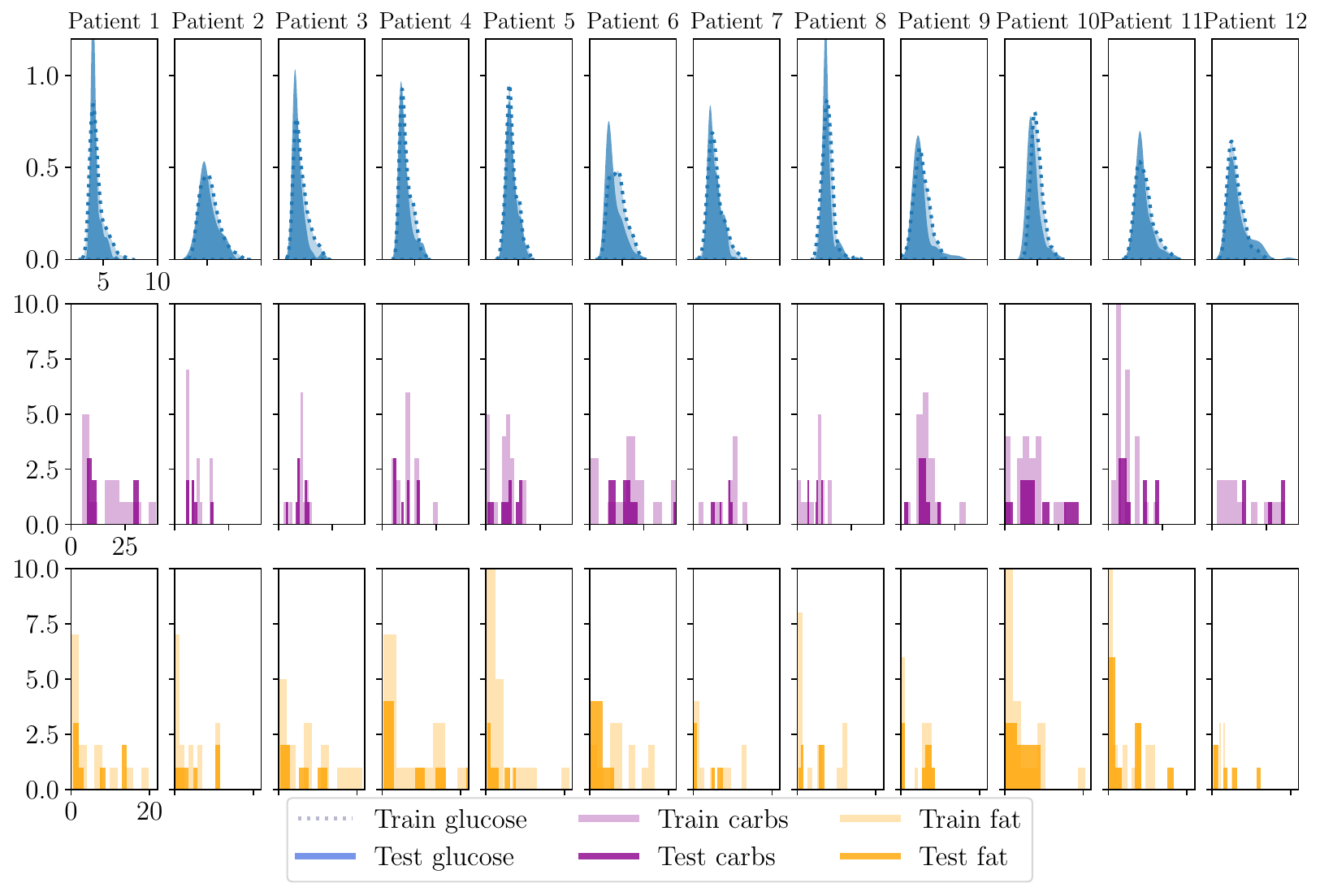}}
\end{figure}

\section{Model training and performance evaluation}\label{apd:sixth}
We illustrate how to assess a model's performance on previously unseen data in \figureref{fig:cv}. The 3-day dataset is partitioned into two segments: the first 2 days and the final 3$^{\text{rd}}$ day. The initial two days are employed for cross-validation and fine-tuning of parameters.

Due to the temporal nature of the data, in each split, the test indices must be temporally later than those in the preceding split. In the $k^{\text{th}}$ split, the first $k$ folds serve as the training set, while the $(k+1)^{\text{th}}$ fold is designated as the validation set.

\begin{figure}[htb]
\floatconts
  {fig:cv}
  {\caption{Process of model parameters tuning and performance evaluation.}}
  {\includegraphics[width=1.0\linewidth, trim={0 0.15cm 0 0}, clip]{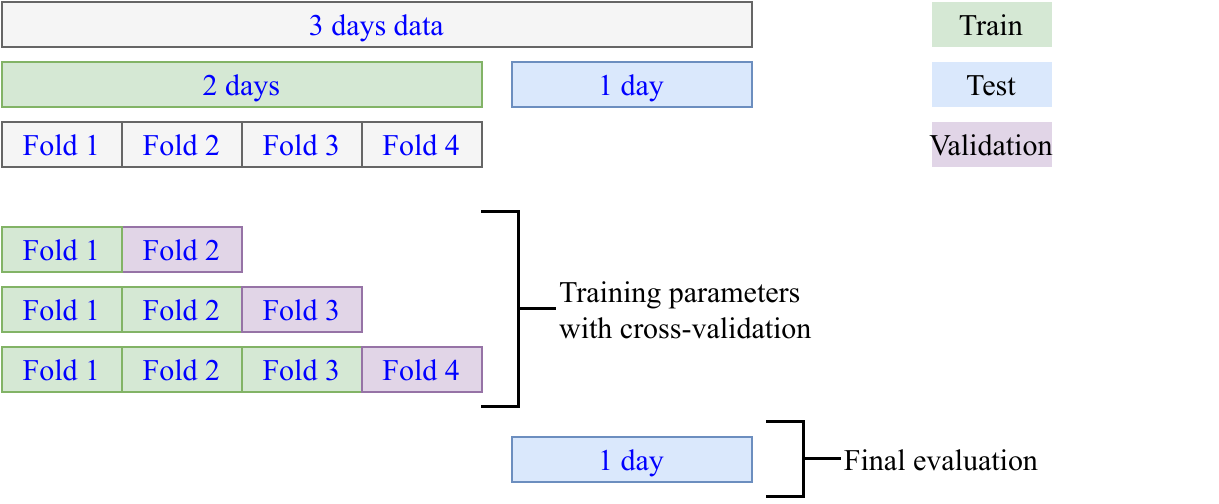}}
\end{figure}

We do not consider cross-validation by patients, as each model possesses parameters tailored to individual patients. Consequently, it cannot be trained and validated solely on specific subsets of patients.

The range of hyperparameter values that we explore during the cross-validation process for the nonparametric models is shown in \tableref{tab:paramscv1}.

We use the best-performing parameter set from the cross-validation to run the model on the one-day test data, thereby obtaining the final metrics shown in \tableref{tab:resultstable}.

\begin{table}[htbp!]
\floatconts
  {tab:paramscv1}%
  {\caption{Hyperparameter search range for nonparametric models}}%
  {%
\resizebox{1.0\columnwidth}{!}{%
\begin{tabular}{lcc}\toprule
Hyperparameter & Search range & Applicable models\\
\midrule
\midrule
Treatment effect time    & 2.5, 3.0, 3.5 & All\\ 
Carbohydrate response lengthscale    & 0.25, 0.3, 0.35, 0.4  & All \\
Fat response lengthscale    & 0.7, 0.75, 0.8, 0.85  &  \modelGPresp, \modelGPlfm\\
Baseline lengthscale & 5, 10, 15 & All \\
\bottomrule
\end{tabular}
}
}
\end{table}

\section{Model predictions}\label{apd:fifth}
The \modelGPlfm's learned latent force and response to the driving component of one patient's meals are shown in \figureref{fig:latents}.
The test set predictions of the non-parametric models are shown in \figureref{fig:predscomp} for several patients. 


\begin{figure}[htb]
\floatconts
  {fig:latents}
  {\caption{Comparison of the \modelGPlfm latent force and final response to carbohydrates for a single patient.}}
  {\includegraphics[width=0.8\linewidth, trim={0 0.15cm 0 0}, clip]{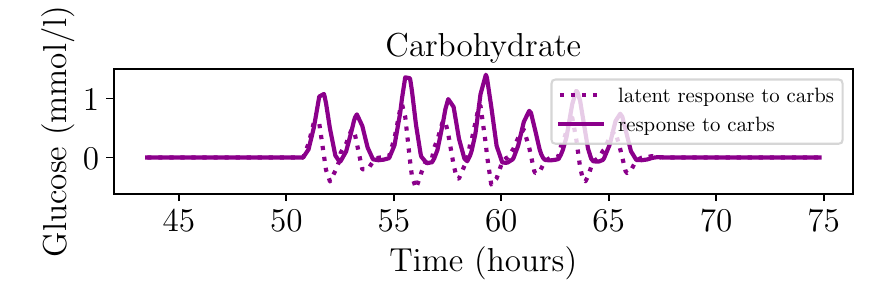}}
\end{figure}

\begin{figure*}[hbt!]
\floatconts
  {fig:predscomp}
  {\caption{Comparison of results of the nonparametric models \modelGPresp, \modelGPlfm, \modelGPconv, \citet{psem_gp_lfm_cheng} and \citet{hizli2023causal}.}}
  {%
    \subfigure[Models' predictions for the 1$^{\text{st}}$ patient][t]{\label{fig:predscomp1}%
      \includegraphics[width=0.7\linewidth]{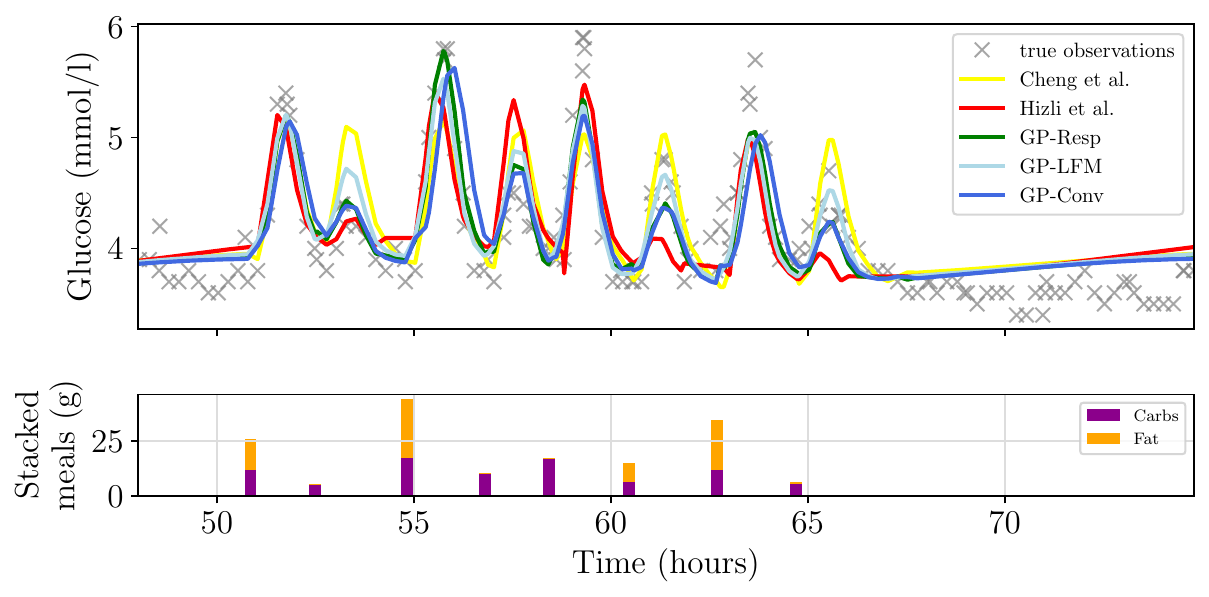}}%
    \qquad
    \subfigure[Models' predictions for the 2$^{\text{nd}}$ patient][b]{\label{fig:predscomp2}%
      \includegraphics[width=0.7\linewidth]{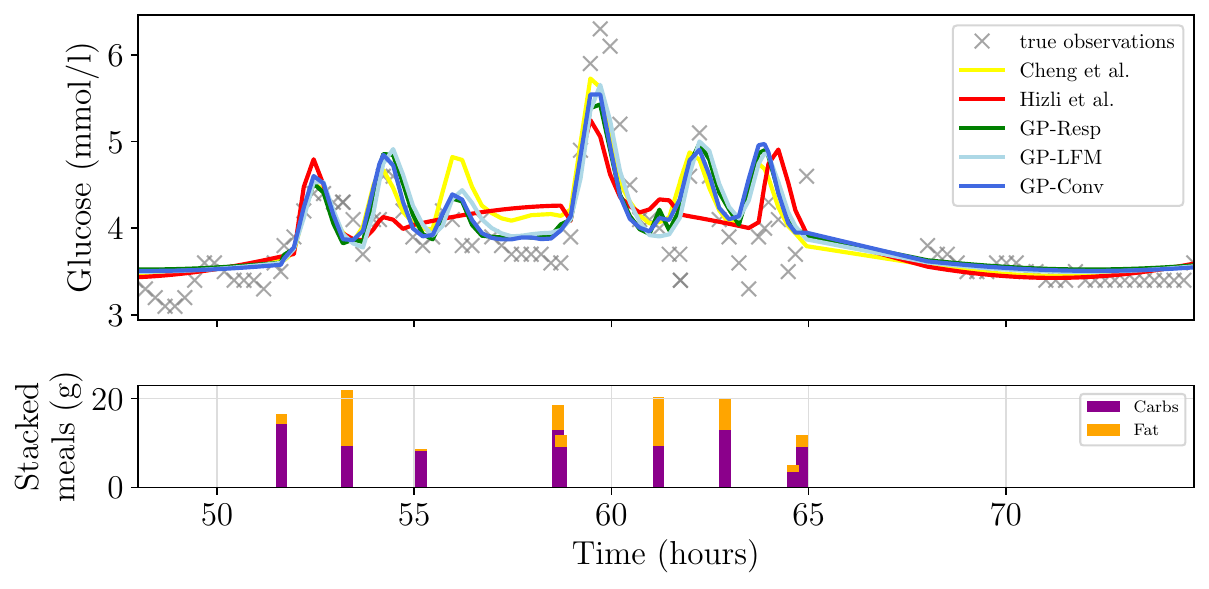}}
          \qquad
    \subfigure[Models' predictions for the 3$^{\text{rd}}$ patient][b]{\label{fig:predscomp3}%
      \includegraphics[width=0.7\linewidth]{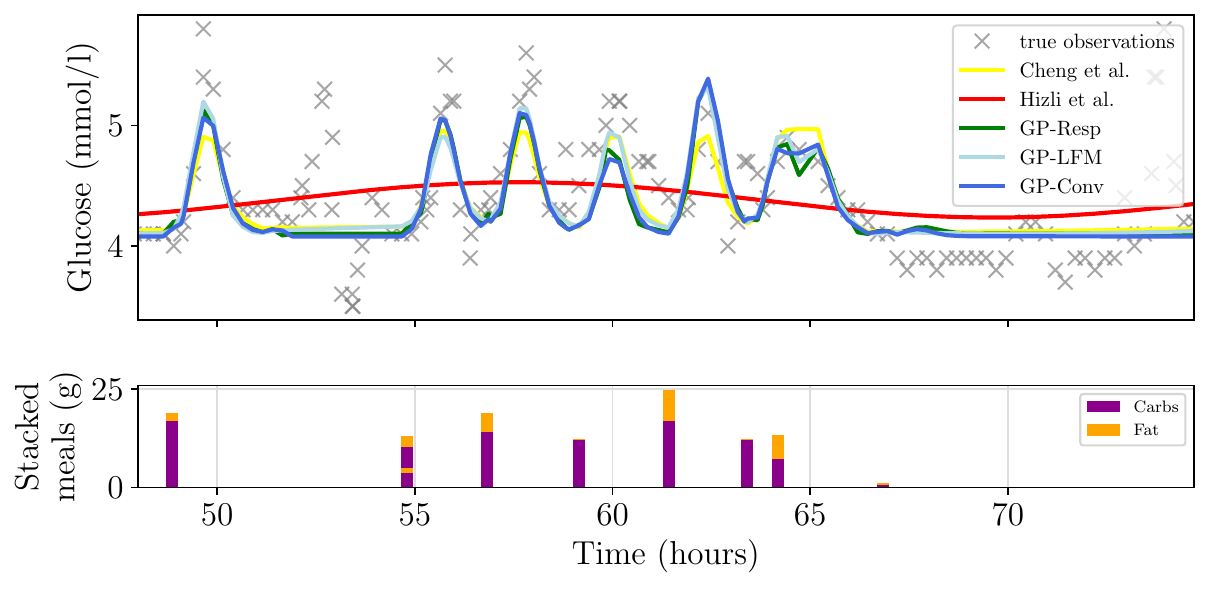}}
  }
\end{figure*}

\end{document}